\pdfoutput=1
\documentclass[conference]{IEEEtran}
\IEEEoverridecommandlockouts
\usepackage{cite}
\usepackage{amsmath,amssymb,amsfonts}
\usepackage{algorithmic}
\usepackage{graphicx}
\usepackage{textcomp}
\usepackage{xcolor}
\usepackage{float}
\usepackage{multicol,subfigure}
\usepackage{cite}
\usepackage{amsmath,amssymb,amsthm,amsfonts,amssymb,bm,mathrsfs}
\usepackage{framed}
\usepackage[linesnumbered,ruled]{algorithm2e} 
\usepackage{algorithmic} 
\usepackage{multirow} 
\usepackage{graphicx}
\usepackage{textcomp}
\usepackage{xcolor,xspace,subfigure,wrapfig}
\usepackage{verbatim}
\usepackage{url}
\usepackage{pst-eps}
\usepackage{pst-plot}
\newcommand{\R}{{\mathbb R}}

\theoremstyle{plain}

\def\BibTeX{{\rm B\kern-.05em{\sc i\kern-.025em b}\kern-.08em
    T\kern-.1667em\lower.7ex\hbox{E}\kern-.125emX}}

\newtheorem{Definition}{\bf Definition}[section]

\newtheorem{Lemma}{\bf Lemma}[section]

\newtheorem{Theorem}{\bf Theorem}[section]

\begin{document}

\title{Accelerated Sparse Bayesian Learning via Screening Test and Its Applications\\
}

\author{\IEEEauthorblockN{Yiping \textsc{JIANG}}
\IEEEauthorblockA{\textit{School of Science and Engineering} \\
\textit{The Chinese University of Hong Kong, Shenzhen}\\
217019004@link.cuhk.edu.cn}
\and
\IEEEauthorblockN{Tianshi \textsc{CHEN}}
\IEEEauthorblockA{\textit{School of Science and Engineering} \\
	\textit{The Chinese University of Hong Kong, Shenzhen}\\
tschen@cuhk.edu.cn}
}

\maketitle

\begin{abstract}
In high-dimensional settings, sparse structures are critical for efficiency in term of memory and computation complexity. For a linear system, to find the sparsest solution provided with an over-complete dictionary of features directly is typically NP-hard, and thus alternative approximate methods should be considered.

In this paper, our choice for alternative method is sparse Bayesian learning, which, as empirical Bayesian approaches, uses a parameterized prior to encourage sparsity in solution, rather than the other methods with fixed priors such as LASSO. Screening test, however, aims at quickly identifying a subset of features whose coefficients are guaranteed to be zero in the optimal solution, and then can be safely removed from the complete dictionary to obtain a smaller, more easily solved problem. Next, we solve the smaller problem, after which the solution of the original problem can be recovered by padding the smaller solution with zeros. The performance of the proposed method will be examined on various data sets and applications.
%

\end{abstract}


\section{Introduction}

For a dynamic system with measurements of input and output signals, system identification is a statistical methodology for building a mathematical model which is powerful enough to describe the characteristics of the system. A classic method for the modeling is called the least squares (LS), to which the systematic treatment is available in many textbooks\cite{Rao}\cite{Smith}\cite{sys_id}. When the LS problems are ill-conditioned, regularization algorithms could be employed to seek optimal solutions. The regularization terms can take various forms, and thus leads to various variants of the regularized least squares. In this thesis, we focus on sparsity inducing regularization.

Finding the sparsest representation of a signal provided with an over-complete dictionary of features is an important problem in many cases, such as signal reconstruction, compressive sensing\cite{app_cs}, feature selection\cite{app_fs}, image restoration\cite{app_ir} and so on. The existing work includes a variety of algorithms. The traditional sparsity inducing regularization methods, including orthogonal matching pursuit (OMP)\cite{omp}, basis pursuit (BP)\cite{BasisPursuit_1994}, LASSO\cite{lasso_1996}, usually prefer a fixed sparsity-inducing prior and perform a standard maximum a posterior probability (MAP)\cite{mle_map} estimation afterwards, thus they can be regarded as Bayesian methods. While in this thesis, we focus on sparse Bayesian learning. This Bayesian method uses a parameterized prior to encourage sparsity, where hyper-parameters are introduced to make the framework more flexible. It's worth mentioning that as an empirical Bayesian method, sparse Bayesian learning has connections with the kernel-based regularization method (KRM)\cite{krm} and machine learning\cite{krm_ml}. When the kernel structure and hyper-parameters are defined specifically, KRM will become sparse Bayesian learning, as discussed in \cite{tac2014}.


In real-world applications, the collected data sets often have large scales and high dimensions, which leads us to consider whether there's a way to screen some features out before solving the high-dimensional problems. We name such operation as ``screening test''. Based on the assumption of sparsity, screening aims to identify features that have zero coefficients and discard them from the optimization safely, therefore the computational burden can be reduced.

In this work, we will propose a screening test for sparse Bayesian learning, and then obtain an accelerated sparse Bayesian learning.

Our contribution can be summarized as follows:

\begin{enumerate}
	\item We propose a screening test for sparse Bayesian learning, which achieves an acceleration in computation time without changing the original optimal solution of the original sparse Bayesian learning.
	\item We examine this accelerated sparse Bayesian learning on various data sets and applications to verify that this method works well for real-world data and problems that can be modeled as linear systems.
\end{enumerate}

And the rest of this note is organized as follows.
In \textit{Section 2}, we introduce sparse Bayesian learning to see its assumptions, methodology, and verify its equivalence to an iterative weighted convex $\ell_1$-minimization problem; in \textit{Section 3}, we design a screening test for the iterative weighted convex $\ell_1$-minimization problem in \textit{Section 2}. The screening test accelerates the computation for each iteration of the $\ell_1$-minimization problem, and thus speeds up the entire sparse Bayesian learning. We check its performance by simulations on two real-world data sets. Next in \textit{Section 4}, we apply the accelerated sparse Bayesian learning method to do classification and verify the trade-off between acceleration and classification accuracy.
In \textit{Section 5}, we use the accelerated sparse Bayesian learning method to do source localization and denoising in astronomical imaging. In this application, not all parameters are linear to the response, thus sampling should be used to deal with the nonlinear ones, then sparse Bayesian learning can play its role. Finally in \textit{Section 6}, we summarize the previous sections.


This work was typeset using \LaTeX. All the simulations were preformed by Python and MATLAB.

\section{Sparse Bayesian Learning}

In this section, we will first introduce linear regression model, and then explore how to find a sparse solution by sparse Bayesian learning (SBL). As the theoretical derivation for SBL has been discussed a lot in \cite{wipf_2003}\cite{wipf_2004}\cite{wipf_2007}, our illustration will be mainly focused on how it can be transformed to a sequence of weighted convex $\ell_1$-minimization problems.

\subsection{Problem Formulation}
The theory of regression aims at modeling relationships among variables and can be used for prediction. Linear regression is an approach to modeling the relationships as linear functions. More specifically, we consider a linear regression model as below:
\begin{equation}\label{eq:lm}
Y={\Phi}\theta+V
\end{equation}
where \({Y\in\R^N}\) is the output, \({{\Phi}}\) =
\([{\phi}_1, {\phi}_2, \ldots, {\phi}_n]\in \R^{N \times n}\)
is the regression matrix made up of $n$ features $\phi_i$, $\phi_i\in\mathbb{R}^N$, \(\theta\in\R^n\) is the parameter to be
estimated, and \(V\in\R^N\) is the noise vector, $V\sim\mathcal{N}(0,\lambda I_N)$, $\lambda\in\R_+$.

One way to estimate $\theta$ is to minimize the least squares (LS) criterion:
\begin{align}
\hat{\theta}^{LS} = \mathop{\arg\min}_{\theta}||Y-\Phi\theta||_2^2=(\Phi^T\Phi)^{-1}\Phi^T Y
\end{align}

When $\Phi\in\R^{N\times n}$ with $N\ll n$ is rank deficient, i.e., $\text{rank}(\Phi)<N$ or close to rank deficient, the LS estimate is said to be ill-conditioned. To handle this issue, the method of regularization could be considered:
\begin{align}
\hat{\theta}^R = \mathop{\arg\min}_{\theta} ||Y-\Phi\theta||_2^2 + \gamma J(\theta) 
\end{align}
where $\gamma\in\R_+$ is called the regularization parameter, and $J(\theta)$ is called the regularization term. There're many choices for $J(\theta)$ with respect to the prior of $\theta$, in this thesis, we focus on sparsity inducing regularization.

Given $\Phi$ and $Y$, to find a sparse \(\theta\), we should solve the
following problem:
\begin{equation}\label{eq:sp}
\begin{split}
\min_{{\theta} \in  { \R } ^ { n }}&\ ||\theta||_0\\
\text{s.t.}&\ ||Y-{\Phi}\theta||_2^2 \leq \epsilon\\
\end{split}
\end{equation}
where \(\epsilon\geq0\) is a tuning parameter that controls the size of
the data fit. The cost function to be minimized represents the
\(\ell_0\) norm of \(\theta\), i.e., the number of non-zero elements in
\(\theta\). Note that problem (\ref{eq:sp}) is combinatorial, which means solving it directly requires an exhaustive search over the entire solution space. For example, in the noise-free case where $V=0$, we have to deal with up to $\binom{n}{N}$ linear systems of size $N\times N$\cite{matrix_computations}. Consequently, approximation methods should be considered. Several approximation methods have been proposed, and one of the most widely used methods is a convex relaxation obtained by replacing the $\ell_0$-norm with the $\ell_1$-norm:
\begin{equation}\label{eq:sp_relax}
\begin{split}
\min_{{\theta} \in  { \R } ^ { n }}&\ ||\theta||_1\\
\text{s.t.}&\ ||Y-{\Phi}\theta||_2^2 \leq \epsilon\\
\end{split}
\end{equation}

The convex relaxation (\ref{eq:sp_relax}) is equivalent to LASSO\cite{lasso_1996}:
\begin{align}\label{lasso}
\min _ {  {\theta} \in  { \R } ^ { n } } \frac { 1 } { 2 } ||   { Y } - \Phi  {\theta} || _ { 2 } ^ { 2 } + \lambda ||  {\theta} || _ { 1 }
\end{align}
where $\lambda\geq 0$ is the regularization parameter.

%
%

\subsection{Methodology}
In this section, we will illustrate the methodology of sparse Bayesian learning (SBL). It was first proposed by Tipping\cite{tipping_2001}, and then applied for signal reconstruction\cite{wipf_2003} and compressive sensing\cite{app_cs}. Compared with classic $\ell_1$-penalty methods like basis pursuit\cite{BasisPursuit_1994} and LASSO\cite{lasso_1996}, sparse Bayesian learning outperforms them in many aspects, for which a reasonable explanation is that one can show SBL is equivalent to an iterative weighted convex $\ell_1$-minimization problem\cite{wipf_2007}.
\subsubsection{Parameterized Prior}
Sparse Bayesian learning\cite{wipf_2003} starts by assuming a Gaussian prior for the parameter $\theta$ as: 
\begin{align}
\theta\sim\mathcal{N}(0,\Gamma(\gamma))
\end{align}
where $\Gamma(\gamma)=\text{diag}(\gamma)$, $\gamma\in\mathbb{R}_{+}^n$. We denote this prior of $\theta$ by $p(\theta;\gamma)$.

Based on the above assumption, sparse Bayesian learning tends to minimize a different cost function in the latent variable space, say $\gamma$-space, where $\gamma\in\R_+^n$ is a vector of $n$ non-negative hyper-parameters governing the prior variance of each unknown $\theta_i$. Since the likelihood $p(Y|\theta)$ is also Gaussian as defined in (\ref{eq:lm}), the corresponding relaxed posterior will be Gaussian:
\begin{align}
p(\theta|Y,\gamma)=\frac{p(Y|\theta)p(\theta;\gamma)}{\int p(Y|\theta)p(\theta;\gamma) d\theta}
\end{align}

Suppose this Gaussian to be $p(\theta|Y,\gamma)=\mathcal{N}(\mu_{\theta},\Sigma_{\theta})$, we can obtain that:
\begin{equation}\label{mu_va_poster}
\begin{split}
\mu_{\theta}=& \Gamma\Phi^T(\lambda I_N +\Phi\Gamma\Phi^T)^{-1}Y \\
\Sigma_{\theta}=& (\Gamma^{-1}+\frac{1}{\lambda}\Phi^T\Phi)^{-1}
\end{split}
\end{equation}
where $\Gamma=\text{diag}(\gamma)$.

\subsubsection{Type-II Estimation}
Mathematically, sparse Bayesian learning tends to select the optimal $\gamma$, say $\hat{\gamma}$, to be the most appropriate $\gamma$ to maximize $p(Y;\gamma)$, which leads to a type-II estimation\cite{mle2}:
\begin{equation}\label{opti_gamma}
\begin{split}
\hat{\gamma} &=\mathop{\arg\max}_{\gamma\succeq 0}  p(Y;\gamma)\\
&=\mathop{\arg\max}_{\gamma\succeq 0}\int p(Y|\theta)p(\theta;\gamma)d\theta\\
&=\mathop{\arg\max}_{\gamma\succeq 0}\int p(Y|\theta) \prod_{i=1}^n \mathcal{N}(\theta_i|0,\gamma_i) d\theta
\end{split}
\end{equation}

Then for the optimal $\hat{\gamma}$, we set a threshold $\epsilon>0$, such that when $\hat{\gamma}_i\leq \epsilon$, the corresponding $\theta_i$ will be $0$.

\begin{Theorem}
	Define $\Sigma_{Y}\triangleq\lambda I_N +\Phi\Gamma\Phi^T$, then it can be proved that the optimal $\hat{\gamma}$ in (\ref{opti_gamma}) can be obtained by minimizing the following function with respect to $\gamma$:
	\begin{align}\label{cost_gamma}
	\text{Loss}(\gamma)=\log|\Sigma_{Y}|+Y^T\Sigma_{Y}^{-1}Y
	\end{align} 
\end{Theorem}

This theorem indicates that we successfully turn the problem (\ref{lasso}) in $\theta$-space into a new problem in $\gamma$-space with respect to the new cost function in (\ref{cost_gamma}).

\subsubsection{Hyper-parameter Estimation}


%
%
%
Since $\log|\Sigma_{Y}|$ is concave in $\gamma$-space, then we can make use of its concave conjugate. Denote $\log|\Sigma_{Y}|$ as $h(\gamma)$, then we have its concave conjugate as:
\begin{align}
h^*( \gamma_h)=\min_{\gamma} { \gamma_h}^T\gamma-h(\gamma)
\end{align}
which indicates that we can also express $h(\gamma)$ as:
\begin{align}
h(\gamma)=\min_{ \gamma_h} { \gamma_h}^T \gamma-h^*( \gamma_h)
\end{align}

Then we obtain an auxiliary cost function for $\text{Loss}(\gamma)$ as:
\begin{align}\label{opt_aux}
\text{Loss}(\gamma, \gamma_h)\triangleq { \gamma_h}^T\gamma-h^*( \gamma_h)+Y^T\Sigma_{Y}^{-1}Y
\end{align}
which should be an upper bound of $\text{Loss}(\gamma)$, i.e.:
\begin{align}
\text{Loss}(\gamma, \gamma_h) \geq \text{Loss}(\gamma)
\end{align}

For any fixed $\gamma$, this bound should be attained by minimizing $\text{Loss}(\gamma, \gamma_h)$ over $ \gamma_h$, indicating that we should choose this optimal value of $ \gamma_h$, denoted by $\gamma_{h_{\text{opt}}}$, as:
\begin{align}\label{opt_dual}
\gamma_{h_{\text{opt}}}=\nabla_{\gamma}(\log|\Sigma_{Y}|)=\text{diag}[\Phi^T\Sigma_{Y}^{-1}\Phi]
\end{align}

Finally, we come to the algorithm for sparse Bayesian learning in \cite{wipf_2007}:

\begin{algorithm}\label{alg_sbl}
	\caption{Sparse Bayesian Learning} 
	Initialize $ \gamma_h$;\footnote{For example, we can initialize $ \gamma_h $ as: $\forall \gamma_{h_i}=1,i=1,\ldots,n$.}
	
	Solve the following optimization problem:
	\begin{align}\label{min_origi}
	\gamma \leftarrow \mathop{\arg\min}_{\gamma}     \text{Loss}_{ \gamma_h}(\gamma) \triangleq  { \gamma_h}^T\gamma+Y^T\Sigma_{Y}^{-1}Y
	\end{align}
	
	Compute the new $ \gamma_h$ based on $\gamma$ according to (\ref{opt_dual});
	
	Repeat (2) and (3), until $\gamma$ converges to some $\gamma_{\text{opt}}$;
	
	Then the optimal $\theta$, denoted by $\theta_{\text{opt}}$, will be obtained as: $\theta_{\text{opt}}= E[\theta|Y;\gamma_{\text{opt}}]=\Gamma_{\text{opt}}\Phi^T(\lambda I_N +\Phi\Gamma_{\text{opt}}\Phi^T)^{-1}Y$.
\end{algorithm}
\bigskip

As for how to find the optimal $\gamma$ in step 2, we have the following lemma from \cite{wipf_2007}:
\begin{Lemma}
	The optimal $\gamma$ in (\ref{min_origi}) can be obtained by solving a weighted convex $\ell_1$-regularized cost function:
	\begin{align}\label{mini_l1}
	\theta^{\text{tmp}}=\mathop{\arg\min}_{\theta}||Y-\Phi\theta||_2^2+2\lambda\sum\limits_{i=1}^n \sqrt{\gamma_{h_i}}|\theta_i| 
	\end{align}
	
	And then we set $\gamma_i=\frac{|\theta_i^{\text{tmp}}|}{\sqrt{\gamma_{h_i}}}, i=1,\ldots,n$. 
\end{Lemma}

%
%
%

By solving a sequence of weighted convex $\ell_1$-minimization problems with respect to $\theta$, we obtain a sparse optimal solution of SBL, where the sparsity is induced by the weighted $\ell_1$ regularization term.

\section{Screening Test for SBL}
\subsection{Motivation}
Screening test aims to quickly identify the inactive features in $\Phi$ that have zero components in the optimal solution $\hat{{\theta}}\ (\text{i.e.}\ \hat{{\theta}}_i=0)$, and then remove them from the optimization without changing the optimal solution. Therefore, the computational cost and memory usage will be saved, especially when $N$ and $n$ are extremely large. For example, when we solve LASSO, the computational complexity of solving it by least angle regression\cite{lasso_sol} is $O(Nn \min\{N, n\})$.

In this section, we will design a screening test for sparse Bayesian learning. Let us first define the index set for the $n$ features $\phi_1,\ldots,\phi_n$ in $\Phi$ as $\mathcal{I}$, i.e. $\mathcal{I}=\{1,2,\ldots,n\}$, then screening test is to find a partition of $\mathcal{I}$ as:
\begin{align}
\mathcal{I}=S \cup \overline{S}, S\cap \overline{S} = \emptyset
\end{align}
where features indexed by $S$ are selected, while the rest features indexed by $\overline{S}$ are rejected.

After the screening, the size of original problem will be reduced. Instead of solving the original problem to obtain the solution $\hat{{\theta}}$ directly, we have an alternative way made up of the three steps below:
\begin{enumerate}
	\item Do the screening to obtain the reduced problem;
	\item Solve the reduced problem to obtain $\hat{\theta}_r$;
	\item Recover $\hat{{\theta}}$ from $\hat{{\theta}}_r$ according to the partition $\mathcal{I}$.
\end{enumerate}

At present, screening rules for LASSO have been explored a lot, which can be roughly divided into two categories: the heuristic screening methods\cite{screen_h1}\cite{screen_h2} and the safe screening methods\cite{screen_s1}\cite{screen_s2}\cite{screen_s3}. The heuristic screening methods, as their name indicates, cannot ensure all the screened features really deserve. In other words, some features that have non-zero coefficients may be mistakenly discarded. However, if the screening is safe, then the reduced problem should be equivalent to the original one. In other words, when all the features indexed in $\overline{S}$ are reasonable to be rejected, the optimal solution $\hat{{\theta}}$ will not change. As for the efficiency of screening, there are two evaluation metrics that we're interested in:

\begin{itemize}
	\item The size of $\overline{S}$ as a fraction of $\mathcal{I}$, say the screening percentage:\\screening percentage$=\frac{\#{\overline{S}}}{\#\mathcal{I}}$.
	\item The total time taken to seek the partition $\mathcal{I}=S \cup \overline{S}$ and to solve the reduced problem relative to the time taken to solve the original problem directly without screening, say the speedup factor:\\
	speedup factor$=\frac{t_{\text{scr}} + t_{\text{red}}} {t_{\text{ori}} }\triangleq \frac{t_r}{t}$.
\end{itemize}
where $t_{\text{ori}}$ is the time to solve the original problem, $t_{\text{scr}}$ is the time to do screening, $t_{\text{red}}$ is the time to solve the reduced problem; and the notation shall be further simplified as $\frac{t_r}{t}$, where $t_r$ is the total time for the reduced case, $t$ is the same as $t_{\text{ori}}$.

\subsection{Methodology}

We try to design a screening test for the optimization problem in line $3$ of Algorithm \ref{alg_sbl}:
\begin{align}\label{lasso_revised}
\min _ {  {\theta} \in  { \R } ^ { n } } \frac { 1 } { 2 } ||   { Y } - \Phi  {\theta} || _ { 2 } ^ { 2 } + \lambda\sum_{i=1}^{n}u_i^{(k)}|\theta_i|
\end{align}
where the second term $\lambda \sum\limits_{i=1}^{n}u_i^{(k)}|\theta_i|$ is a weighted $\ell_1$-norm of $\theta$.

This is a LASSO-type problem. Based on the screening tests for LASSO\cite{Xiang2011}\cite{wangjie}\cite{XiangWR14}, we will propose a safe screening test for (\ref{lasso_revised}), where the procedure including models, theorems, lemmas and so on, must be revised accordingly. To guarantee the accuracy and completeness of the thesis, we will go through all the details including proofs during the revision. Let's start from the dual formulation first.

\subsubsection{Dual Formulation}

By introducing \(z=Y-\Phi\theta\) into (\ref{lasso_revised}), the primal problem becomes:
\begin{align}\label{primal_z}
\begin{split}
\min_{\theta\in\R^n}\ &\frac{1}{2}||z||_2^2+\lambda\sum_{i=1}^{n}u_i^{(k)}|\theta_i|\\
\text{s.t.}\ &z=Y-\Phi\theta
\end{split}
\end{align}

Moreover, it can be proved that the dual problem of (\ref{lasso_revised}) should be:
\begin{align}\label{dual_u}
\begin{split}
\max_{\eta\in\R^N}\ &\frac{1}{2}||Y||^2-\frac{1}{2}||\eta-Y||^2\\
\text{s.t.}\ &| {\frac{\phi_i^T\eta}{\lambda u_i^{(k)}} } |\leq 1,i=1,\ldots,n
\end{split}
\end{align}

Note that (\ref{primal_z}) is a convex problem with affine constraints. By Slater's
condition\cite{convex_opt}, as long as the problem is feasible, strong duality will hold. Then we denote \((\hat{\theta},\hat{z})\) and
\(\hat{\eta}\) as optimal primal and dual variables,
and make use of the Lagrangian again:
\begin{align}
\mathcal{L}(\theta,z,\eta)=\frac{1}{2}||z||_2^2+\lambda \sum_{i=1}^{n}u_i^{(k)}|\theta_i|+\eta^T(Y-\Phi\theta-z)
\end{align}

According to the Karush–Kuhn–Tucker (KKT) conditions\footnote{Here we use the subgradient for $\theta$ because $\ell_1$ norm is not differentiable at the kink.}, we have:
\begin{align}
\begin{split}
_{\theta}0\in\partial_{\theta} \mathcal{L}(\hat{\theta},\hat{z},\hat{\eta})&=-\Phi^T\hat{\eta}+\lambda u^{(k)}.*v,\\
&\text{where}\ ||{v}||_{\infty}\leq 1\ \text{and}\ {v}^T\hat{\theta}=||\hat{\theta}||_1;\\
\nabla_z \mathcal{L}(\hat{\theta},\hat{z},\hat{\eta}) &= \hat{z}-\hat{\eta}=0;\\
\nabla_{\eta} \mathcal{L}(\hat{\theta},\hat{z},\hat{\eta}) &= Y-\Phi\hat{\theta}-\hat{z}=0.
\end{split}
\end{align}

By solving the equations above, we have:
\begin{align}
Y=\Phi\hat{\theta}+\hat{\eta}
\end{align}

And there exists a specific \(\hat{v}\in\partial||\hat{\theta}||_1\) such that:
\begin{equation}
u.*\hat{v}=\frac{{\Phi^T}{\hat{\eta}}}{\lambda},||\hat{v}||_{\infty}\leq1,\hat{v}^T\hat{\theta}=||\hat{\theta}||_1
\end{equation}
which is equivalent to:
\begin{equation}
|\frac{\phi_i^T\hat{\eta}}{ \lambda u_i^{(k)} }|\leq 1,i=1,2,...n
\end{equation}

And we can further conclude:
\begin{equation}\label{nec_condi}
\begin{split}
\frac{\hat{\eta}^T\phi_i}{ \lambda u_i^{(k)} }=\left\{\begin{array}{ll}{\text{sign}(\hat{\theta}_i),} & {\text { if } \hat{\theta}_{i}\neq 0} \\ {[-1,1],}  & {\text { if } \hat{\theta}_{i}=0}\end{array}\right.
\end{split}
\end{equation}
which indicates the theorem below:

\begin{Theorem}\label{thm:suff}
	\begin{equation}
	|\frac{\hat{\eta}^T\phi_i}{\lambda u_i^{(k)}}|<1 \Rightarrow \theta_i=0, i=1,\ldots,n
	\end{equation}
\end{Theorem}

\subsubsection{Region Test}

Theorem \ref{thm:suff} works as a sufficient condition to reject $\phi_i$: 
\begin{align}
|\frac{\hat{\eta}^T\phi_i}{\lambda u_i^{(k)}}|<1
\end{align}

i.e.:
\begin{align}
\max\{\frac{\hat{\eta}^T\phi_i}{\lambda u_i^{(k)}},-\frac{\hat{\eta}^T\phi_i}{\lambda u_i^{(k)}}\}< 1
\end{align}

However, the optimal $\hat{\eta}$ is not available, which leads us to consider alternative methods. Region test is a good choice which works by bounding $\hat{\eta}$ in a region $\mathcal{R}$. Since there might be vectors other than $\hat{\eta}$ in $\mathcal{R}$, it will be harder\footnote{Even if $\phi_i$ can make it to satisfy the sufficient condition, other vectors will possibly fail the condition.} for us to reject each $\phi_i$, therefore the sufficient condition will be relaxed. This relaxation can be expressed as a new theorem:

\begin{Theorem}
	Suppose we find a region $\mathcal{R}$ such that $\hat{\eta}\in\mathcal{R}$, then:
	\begin{equation}
	|\frac{{\eta}^T\phi_i}{\lambda u_i^{(k)}}|<1, \forall \eta\in\mathcal{R} \Rightarrow \theta_i=0, i=1,\ldots,n
	\end{equation}
\end{Theorem}

Note that the optimal $\mathcal{R}$ in theory should be $\mathcal{R}=\{\hat{\eta}\}$. For convenience, we will define $\mu_{\mathcal{R}}(\phi_i)=\max\limits_{\eta\in\mathcal{R}}  \frac{{\eta}^T\phi_i}{\lambda u_i^{(k)}}$, then the sufficient condition will become:
\begin{align}\label{criterion}
\max \{\mu_{\mathcal{R}}(\phi_i), \mu_{\mathcal{R}}(-\phi_i)\} <1 \Rightarrow \theta_i=0, i=1,\ldots,n
\end{align}

Next, we will try to find an appropriate region $\mathcal{R}$. For the design of the region, the idea is quite similar to that of \cite{Xiang2011} and \cite{XiangWR14}, however to guarantee the accuracy and completeness of the thesis, we will go through the construction of $\mathcal{R}$ from scratch.

\noindent{\textbf{Sphere Test}}

The simplest region is a sphere\cite{Xiang2011} decided by observing the objective function in (\ref{dual_u}). We notice that $\eta$, as the dual variable of $\theta$, is the projection of $Y$ on the feasible set: 
\begin{equation}
\mathcal{F}=\{ \eta:| {\frac{\phi_i^T\eta}{\lambda u_i^{(k)}} } |\leq 1,i=1,\ldots,n \}
\end{equation}

If we can find a feasible point $\eta '$, then we will obtain a sphere to bound $\hat{\eta}$ with $\eta'$ as the sphere center. The sphere center can be chosen as:
\begin{align}
\eta'=\frac{\lambda Y u_\text{min}^{(k)}}{\lambda_\text{max}}
\end{align}
where $\lambda_\text{max}=\max\limits_{i} \phi_i^T Y$, $u_\text{min}^{(k)}=\min \{ u_1^{(k)},\ldots, u_n^{(k)} \}$. Then the sphere should be:
\begin{align}\label{sphere}
B(c,r)=\{\eta:||\eta-c||_2\leq r\}
\end{align}
where $c=Y$, $r=||\eta'-Y||_2$. We can further compute the corresponding $\mu_{B}(\phi_i)$ as below:
\begin{align}
\mu_{B}(\phi_i) &= \max\limits_{\eta\in B(c,r)}\frac{\eta^T\phi_i}{\lambda u_i^{(k)}}\\
&=\frac{1}{\lambda u_i^{(k)}}(\eta^T-c^T)\phi_i+\frac{c^T}{\lambda u_i^{(k)}}\phi_i\\
&\leq \frac{r}{\lambda u_i^{(k)}}||\phi_i||_2+\frac{c^T\phi_i}{\lambda u_i^{(k)}}
\end{align}

Thus according to (\ref{criterion}), the sphere test should be:
\begin{align}\label{sphere_test_u}
T_{B}(\phi_i)=\left\{\begin{array}{ll}{1,} & {\text { if }\left|c^{T} \phi_i\right|<\lambda u_i^{(k)}-r||\phi_i||_{2}} \\ {0,} & {\text { otherwise. }}\end{array}\right.
\end{align}
where $1$ indicates $\theta_i$ is zero and $\phi_i$ can be rejected, $0$ indicates $\theta_i$ is non-zero and $\phi_i$ should be reserved.

\noindent{\textbf{Dome Test}}

Based on sphere test, we improve the region that bounds the optimal $\hat{\eta}$ by introducing a hyperplane\cite{Xiang2011}, then the new region should be defined as:
$$ \mathcal{D}(c,r;n,h) = \{\eta:{n}^{T} {\eta} \leq h, ||{\eta}-c||_{2} \leq r\} $$
which shares the same $c,r$ as sphere test, however will further select a specific pair of $(n,h)$ among the $2n$ linear constraints (half spaces) in (\ref{dual_u}), where $n = \pm \frac{\phi_i}{||\phi_i||_2}$, $h=\frac{\lambda u_i^{(k)}}{||\phi_2||_2}$. With proper selection of $n$ and $h$, the selected hyperplane will cut into the sphere and bound the optimal $\hat{\eta}$ in a tighter region. We call such a region as ``dome''.

For the selection of $n$ and $h$, we should define the following variables as preparations:

\begin{itemize}
	\item $c_d$, the dome center on the hyperplane,  for which the line passing $c$ and itself is in the direction $-n$;
	\item $\psi_d$, the fraction of the signed distance from $c$ to $c_d$ compared with the sphere radius $r$;
	\item $r_d$, the largest distance a point can move from $c_d$ within the dome and hyperplane.
\end{itemize}

These variables can be expressed in geometry as:


\begin{figure}[H]\label{region}
	\centering
	\includegraphics[width=0.4\linewidth]{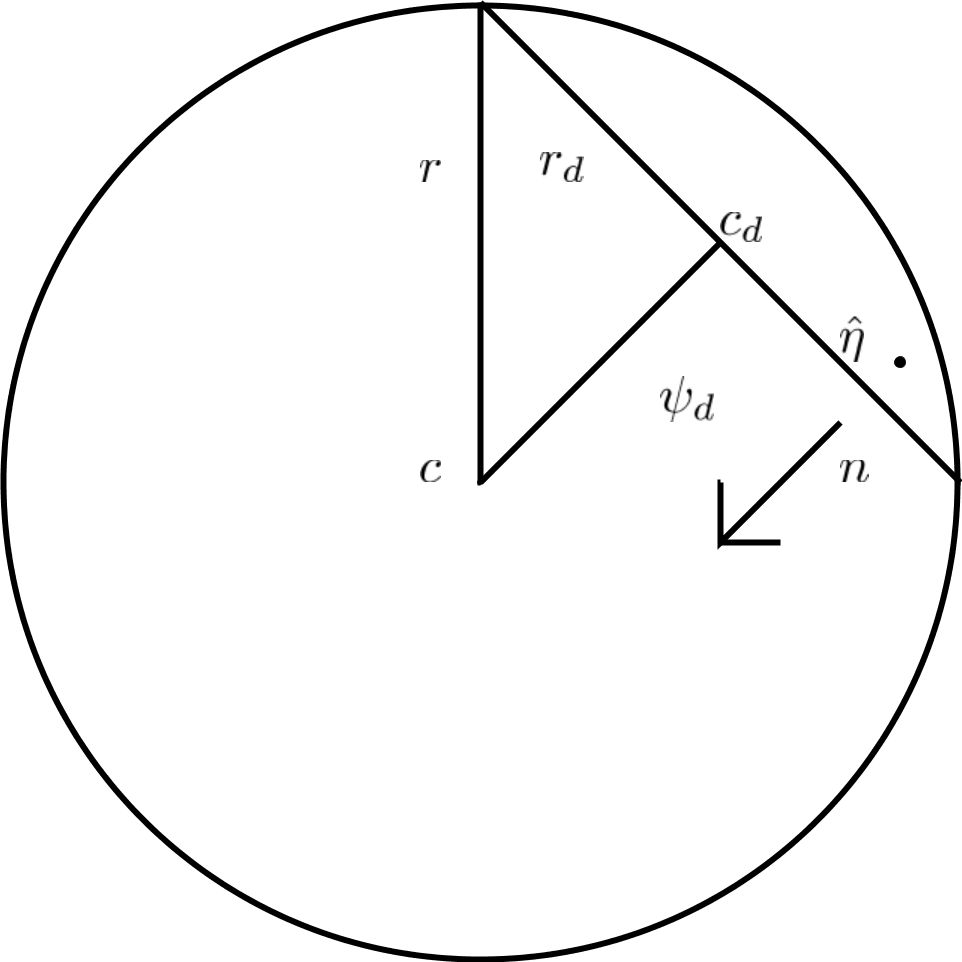}
	\caption{Dome test}
	\label{fig:4}
\end{figure}

By Euclidean geometry, the following relationships among these variables will be obtained:
\begin{align}
\begin{array}{c}{\psi_{d}=\left({n}^{T} c-h\right) / r} \\ {{c}_{d} = {c}-\psi_{d} r {n}} \\ {r_{d}=r \sqrt{1-\psi_{d}^{2}}}\end{array}
\end{align}

To ensure that $c_d$ is inside the sphere $B(c,r)$, we require $-1\leq\psi_{d}\leq 1$. Now, the optimal $\phi$, say $\phi_g$, should be the $\pm \phi_i$ that attains the smallest intersection of one half space and the sphere, thus $\psi_{d}$ should be maximized:
\begin{align}\label{phig_u}
\phi_g=\mathop{\arg\max}_{\{\pm\phi_i\}_{i=1}^n} \frac{ \phi_i^{T} c-\lambda u_i^{(k)}}{||\phi_i||_{2}}
\end{align}

This optimal $i$ will be recorded as $i^*$ in discussion afterwards. As for $\mu(\phi_i)$ for dome, we have the following lemma revised from \cite{Xiang2011}:

\begin{Lemma}\label{lem:dt}
	For a fixed dome $\mathcal{D}(c,r;n,h)$ satisfying $|\psi_{d}|\leq 1$, the corresponding $\mu_{\mathcal{D}}(\phi_i)$ should be:
	\begin{align}\label{DT_criterion}
	\mu_{\mathcal{D}}(\phi_i) =  \frac{1}{\lambda u_i^{(k)}} [c^T\phi_i+M_1(n^T\phi_i,||\phi_i||_2)]
	\end{align}
	where
	\[
	M_1(t_1,t_2) = \\
	\left\{\begin{array}{ll}{rt_2,}  {\text { if } t_1<-\psi_dt_2} \\
	{-\psi_{d}rt_1+r\sqrt{t_2^2-t_1^2}\sqrt{1-\psi_d^2},}  {\text { if } t_1\geq-\psi_dt_2}\\
	\end{array}\right.
	\] 
\end{Lemma}

Thus the dome test should be designed as:
\begin{Theorem}\label{thm:dt}
	The screening test for a fixed dome $\mathcal{D}(c,r;n,h)$ should be:
	\begin{align}
	T_{\mathcal{D}}(\phi_i) = \left\{
	\begin{array}{ll}{1,} & { \text { if } V_l(n^T\phi_i,||\phi_i||_2)< c^T\phi_i <V_u(n^T\phi_i,||\phi_i||_2) } \\
	{0,} & {\text { otherwise. }}\end{array}\right.
	\end{align}
	where $V_u(t_1,t_2)=\lambda u_i^{(k)}-M_1(t_1,t_2)$ and $V_l(t_1,t_2)=-V_u(-t_1,t_2)$.
\end{Theorem}
%
%

\noindent{\textbf{Two Hyperplane Test}}

Based on the dome test, we try to introduce one more hyperplane to the region\cite{XiangWR14}, which ensures a better bound for $\hat{\eta}$. However, it's necessary to guarantee that the new intersection of a single sphere and two hyperplanes should be non-empty. For this purpose, we make use of the following lemma from \cite{XiangWR14}:

\begin{Lemma}
	Let the sphere $B(c,r)$ and half space $(n,h)$ bound the dual optimal solution $\hat{{\eta}}$ with the dome $ \mathcal{D}(c,r;n,h) = \{\eta:{n}^{T} {\eta} \leq h, ||{\eta}-c||_{2} \leq r\} $ satisfying $0<\psi_{d}\leq 1$, then the new sphere $B(c_d,r_d)$, which is smaller than the original $B(c,r)$, is the circumsphere of the dome $\mathcal{D}$ and thus still bounds $\hat{{\eta}}$.
\end{Lemma}

Based on this lemma, we can name $\phi_g$ as $\phi^{(1)}$, and then select a $\phi^{(2)}$ other than $\phi^{(1)}$. This $\phi^{(2)}$ should ensure the smallest intersection of $B(c_d,r_d)$ and one of the rest half spaces:
\begin{align}\label{phi2_u}
\phi^{(2)}=\mathop{\arg\max}_{\{\pm\phi_i\}_{i=1}^n \backslash\phi^{(1)}} \frac{ \phi_i^{T} c_d-\lambda u_i^{(k)}}{||\phi_i||_{2}}
\end{align}

We call this optimal $i(i\neq i^*)$ as $j^*$, and now the intersection among one sphere and two hyperplanes should be non-empty. Therefore we can finally define the region denoted by $\mathcal{H}_2$ as:
\begin{align}
\mathcal{H}_2 = \mathcal{H}_2(c,r;n_1,h_1,n_2,h_2)
\end{align}
where $c=Y$, $r=||\eta'-Y||_2$, $n_1 = \frac{\phi^{(1)}}{||\phi^{(1)}||_2}$, $h_1=\frac{\lambda u_{i^*}^{(k)}}{||\phi^{(1)}||_2}$, $n_2 = \frac{\phi^{(2)}}{||\phi^{(2)}||_2}$, $h_2=\frac{\lambda u_{j^*}^{(k)}}{||\phi^{(2)}||_2}$. And we can express the region as the figure below:

\begin{figure}[H]\label{region_tht}
	\centering
	\includegraphics[width=0.6\linewidth]{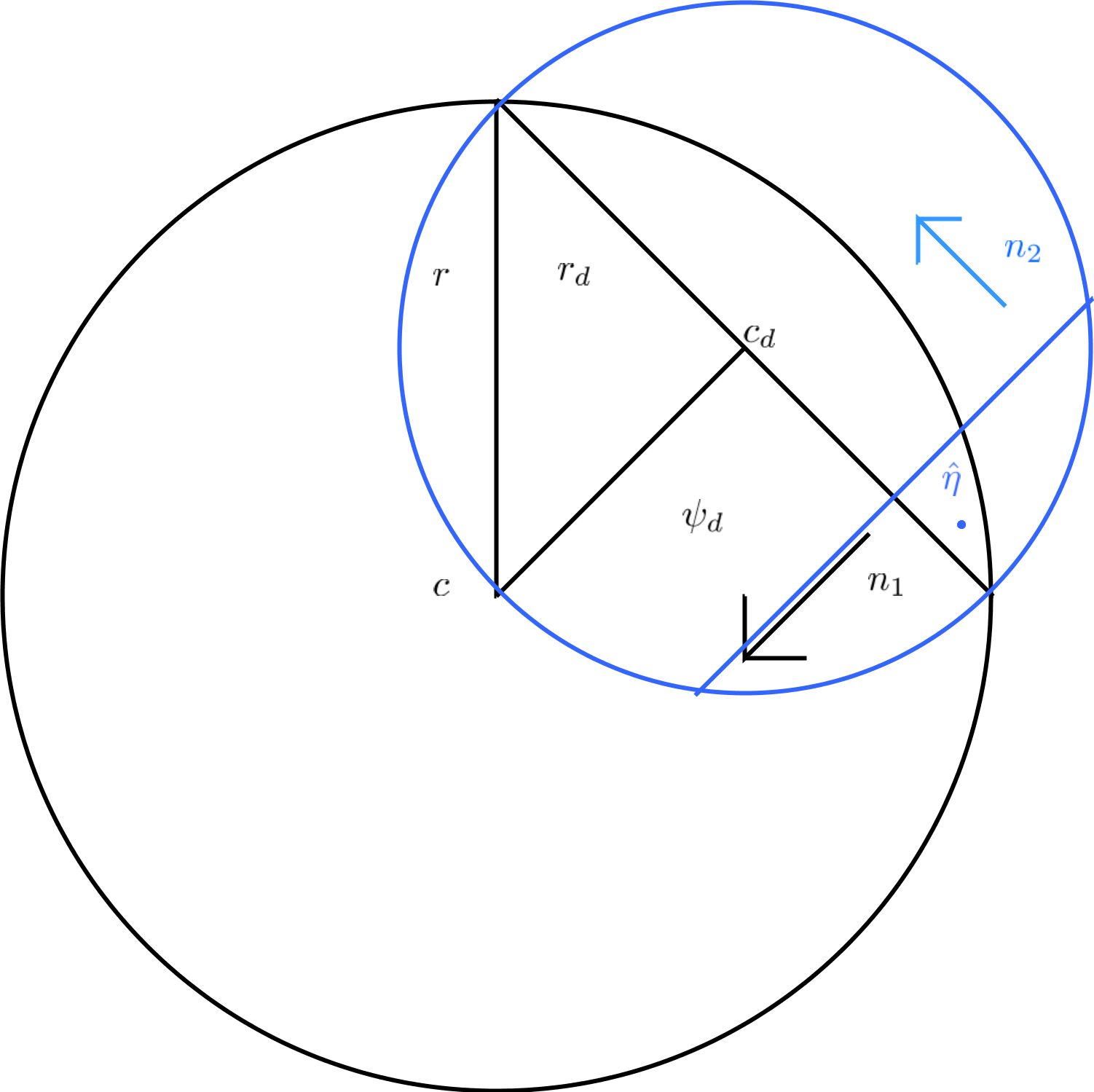}
	\caption{Two hyperplane test}
	\label{fig:5}
\end{figure}

As for the criterion $\mu(\phi_i)$ for $\mathcal{H}_2$, we revise the lemma in \cite{XiangWR14} to obtain:

\begin{Lemma}
	
	Fix the region $\mathcal{H}_2(c,r;n_1,h_1;n_2,h_2)$, suppose $\psi_i$ satisfies $\left|\psi_i\right|\leq1,i=1,2$ and $\arccos \psi_1 + \arccos \psi_2\geq \arccos (n_1^T n_2)$, define:
	\begin{align}
	h(x,y,z)=\sqrt{ (1-\tau^2)z^2+2\tau xy-x^2-y^2 }
	\end{align}
	where $\tau = n_1^T n_2$. Then for $\phi_i\in \R^N$, we have:
	\begin{align}\label{THT_criterion}
	\mu_{\mathcal{H}_2}(\phi_i) =\frac{1}{\lambda u_i^{(k)}}[ c^T\phi_i+M_2(n_1^T\phi_i,n_2^T\phi_i,||\phi_i||_2)]
	\end{align}
	where\\
	\[
	M_2(t_1,t_2,t_3) = \\
	\left\{\begin{array}{ll}{rt_3,}  {\text { if } (a)} \\
	{-rt_1\psi_1+r\sqrt{t_3^2-t_1^2}\sqrt{1-\psi_1^2},}  {\text { if } (b)}\\
	{-rt_2\psi_2+r\sqrt{t_3^2-t_2^2}\sqrt{1-\psi_2^2},}  {\text { if } (c)}\\
	{ -\frac{r}{1-r^2}[(\psi_1-\tau\psi_2)t_1+(\psi_2-\tau\psi_1)t_2] }\\
	{+\frac{r}{1-r^2}h(\psi_1,\psi_2,1)h(t_1,t_2,t_3), }\  {{otherwise}} \end{array}\right.
	\]	
	and conditions $(a),(b),(c)$ are given by:
	\begin{align}
	(a)&\ t_1<-\psi_1t_3\ \&\ t_2<-\psi_2t_3;\\
	(b)&\ t_1\geq-\psi_1t_3\ \&\  \frac{\left(t_{2}-\tau t_{1}\right)} {\sqrt{t_{3}^{2}-t_{1}^{2}}} < \frac{\left(-\psi_{2}+\tau \psi_{1}\right)} { \sqrt{1-\psi_{1}^{2}}};\\	
	(c)&\ t_2\geq-\psi_2t_3\ \&\  \frac{\left(t_{1}-\tau t_{2}\right)} {\sqrt{t_{3}^{2}-t_{2}^{2}}} < \frac{\left(-\psi_{1}+\tau \psi_{2}\right)} {\sqrt{1-\psi_{2}^{2}}};
	\end{align}
\end{Lemma}

Then the two hyperplane test can be designed as:
\begin{Theorem}
	
	The two hyperplane test for the region $\mathcal{H}_2 = \mathcal{H}_2(c,r;n_1,h_1;n_2,h_2)$ is:
	\[
	T_{\mathcal{H}_2}(\phi_i) = \left\{\begin{array}{ll}{1,} & {\text { if } (a')} \\ {0,} & {\text { otherwise. }}\end{array}\right.
	\]
	where condition $(a')$ is:
	$$V_{l}\left({n}_{1}^{T} \phi_i, {n}_{2}^{T} \phi_i,||\phi_i||_{2}\right)<c^{T} \phi_i<V_{u}\left({n}_{1}^{T} \phi_i, {n}_{2}^{T} \phi_i,||\phi_i||_{2}\right)$$
	with $V_u(t_1,t_2,t_3)=\lambda u_i^{(k)}-M_2(t_1,t_2,t_3)$ and $V_l(t_1,t_2,t_3)=-V_u(-t_1,-t_2,t_3)$.
\end{Theorem}

%

As for the computational complexity, the two hyperplane test requires $n$ triples of $(n_1^T\phi_i,n_2^T\phi_i,||\phi_i||_2)$ with the help of $u_i^{(k)}$, $i=1,2,\ldots,n$, thus the computational complexity should be $O(Nn)$. It's also worth mentioning that if we continue increasing the number of hyperplanes, the region test should be more complicated, however will have the potential to reject more features since the region that bounds $\hat{\eta}$ should be tighter. Here we stop at $m=2$, and summarize the new algorithm which is similar to the THT algorithm in \cite{XiangWR14} as Algorithm \ref{alg_wtht}, and name it as weighted-THT (W-THT):

\begin{algorithm}\label{alg_wtht}
	\caption{Weighted Two Hyperplane Test} 
	\KwIn{$Y,\lambda,\Phi=\{\phi_1,\dots,\phi_n\},u^{(k)} $.}
	\KwOut{$v=\{v_1,v_2,\dots,v_n\}$. (If $v_i=1$, then $\phi_i$ is rejected)}
	
	$\phi^{\text{norm}}_i\leftarrow ||\phi_{i}||_2,i=1,\dots,n$;
	
	$c\leftarrow Y$;(sphere center)
	
	$\rho_i\leftarrow c^T\phi_{i},i=1,\dots,n$;
	
	$\lambda_{\text{max}}\leftarrow {\text{max}_i\left|\rho_i\right|}$;
	
	$\eta_{\mathcal{F}}=\frac {\lambda Y u_\text{min}^{(k)}} {\lambda_{\text{max}}}$;
	
	$r\leftarrow||\eta_{\mathcal{F}}-c||_2$;(sphere radius)
	
	$i_*\leftarrow \arg\max_i\frac{\left|\rho_i\right|-\lambda u_i^{(k)}}{\phi^{\text{norm}}_i}$;
	
	$n_1\leftarrow \text{sign}(\rho_{i^*})\phi_{i_*}/{\phi^{\text{norm}}_{i_*}}$;
	
	$h_1\leftarrow \lambda u_{i^*}^{(k)}/{\phi^{\text{norm}}_{i_*}}$;
	
	$a\leftarrow n_1^Tc-h_1$;
	
	$\sigma_i\leftarrow n_1^T\phi_i,i=1,\ldots,n$;
	
	$t_i\leftarrow \rho_i-a\sigma_i,i=1,\ldots,n$;
	
	$j_*\leftarrow \arg\max \limits_{i\neq i_*}\frac{\left|t_i\right|-\lambda u_i^{(k)}}{\phi^{\text{norm}}_i}$;
	
	$n_2\leftarrow \text{sign}(\rho_{j^*})\phi_{j_*}/{\phi^{\text{norm}}_{j_*}}$;
	
	$h_2\leftarrow \lambda u_{j^*}^{(k)} /{\phi^{\text{norm}}_{j_*}}$;
	
	$\tau_i \leftarrow n_2^T\phi_{i}, i=1,\ldots,n$;
	
	$v_i\leftarrow [V_l(\sigma_i,\tau_i,{\phi^{\text{norm}}_i})<\rho_i<V_u(\sigma_i,\tau_i,{\phi^{\text{norm}}_i})].$
\end{algorithm}

In line $17$: for condition $a$, $[a]$ returns to $1$ (TRUE) if $x$ is true.

\subsection{Simulation}

In this section, we conduct experiments to verify that the proposed sparse Bayesian learning with screening test does outperform in speed while keeping the optimal solution unchanged at the same time. To solve (\ref{lasso_revised}) we use CVX, a package for specifying and solving convex programs\cite{cvx_1}\cite{cvx_2}.

\subsubsection{Real-world Data Sets}
The experiments are based on real-world data sets. These data sets often have complicated structures which will affect the performance of the screening, and we will model them as a linear system in (\ref{eq:lm}). The two data sets we used are listed as below:

\begin{itemize}
	\item MNIST handwritten image data (MNIST)\cite{mnist1}.
	
	MNIST is made up of $70,000$ images ($28\times 28$) as a record for handwritten digits. It has $60,000$ images in the training set and $10,000$ images in the testing set. We will vectorize all the images as $784$-dimensional vectors and scale them to unit norm. Then we randomly selected $10,000$ images in the training set to be the columns of our regression matrix $\Phi$ ($1,000$ for each digit), and randomly sample one target image from the testing set as $Y$. Therefore, we will do a simulation with $\Phi\in\R^{784\times 10000}$ and $Y\in\mathbb{R}^{784}$.
	
	\item New York Times bag-of-words data (NYTW)\cite{Lichman}.
	
This data set can be downloaded from the UCI Machine Learning Repository. The raw data can be stored as a matrix which contains $300,000$ documents expressed as vectors with respect to a vocabulary of $102,660$ words. In this matrix, the element $(i,j)$ represents the number of occurrences of the $i$th word in the $j$th document. We will preprocess the raw data by randomly selecting $50,000$ documents and $5,000$ words to become the regression matrix $\Phi\in\R^{5000\times 50000}$; and the response $Y\in\R^{5000}$ will be the subset of randomly-chosen document column with respect to the $5,000$ words in the regression matrix.
	
\end{itemize}

\subsubsection{Results and Analysis}

When it comes to the performance of the proposed method, we should set a metric for different data sets. A possible choice is to make use of $\lambda_{\text{max}}$. Recall that we define $\lambda_{\text{max}}=\max\limits_{i} \phi_i^T Y$ during the construction of sphere, then we can use the ratio $\lambda/\lambda_{\text{max}}$ as measure of regularization. The simulation results for MNIST with respect to screening percentage and time reduction are shown as the following two figures:

%
%

\begin{figure}[H]
	\centering
	\includegraphics[width=0.6\linewidth]{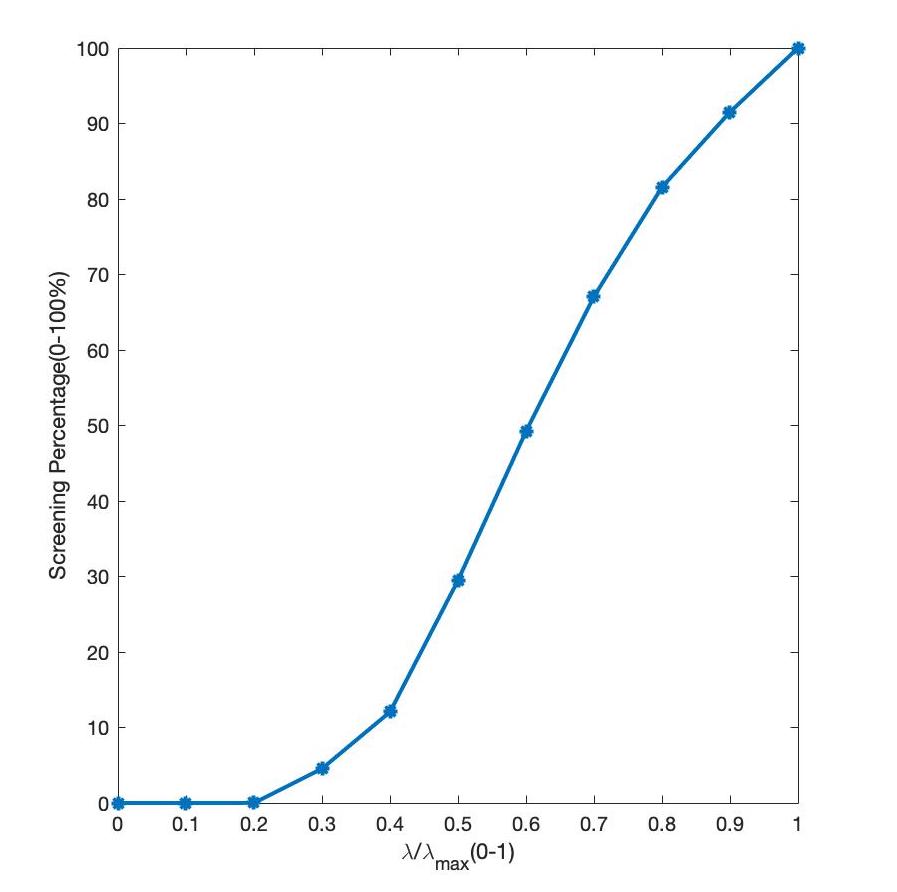}
	\caption{Screening percentage - MNIST}
\end{figure}

\begin{figure}[H]
	\centering
	\includegraphics[width=0.6\linewidth]{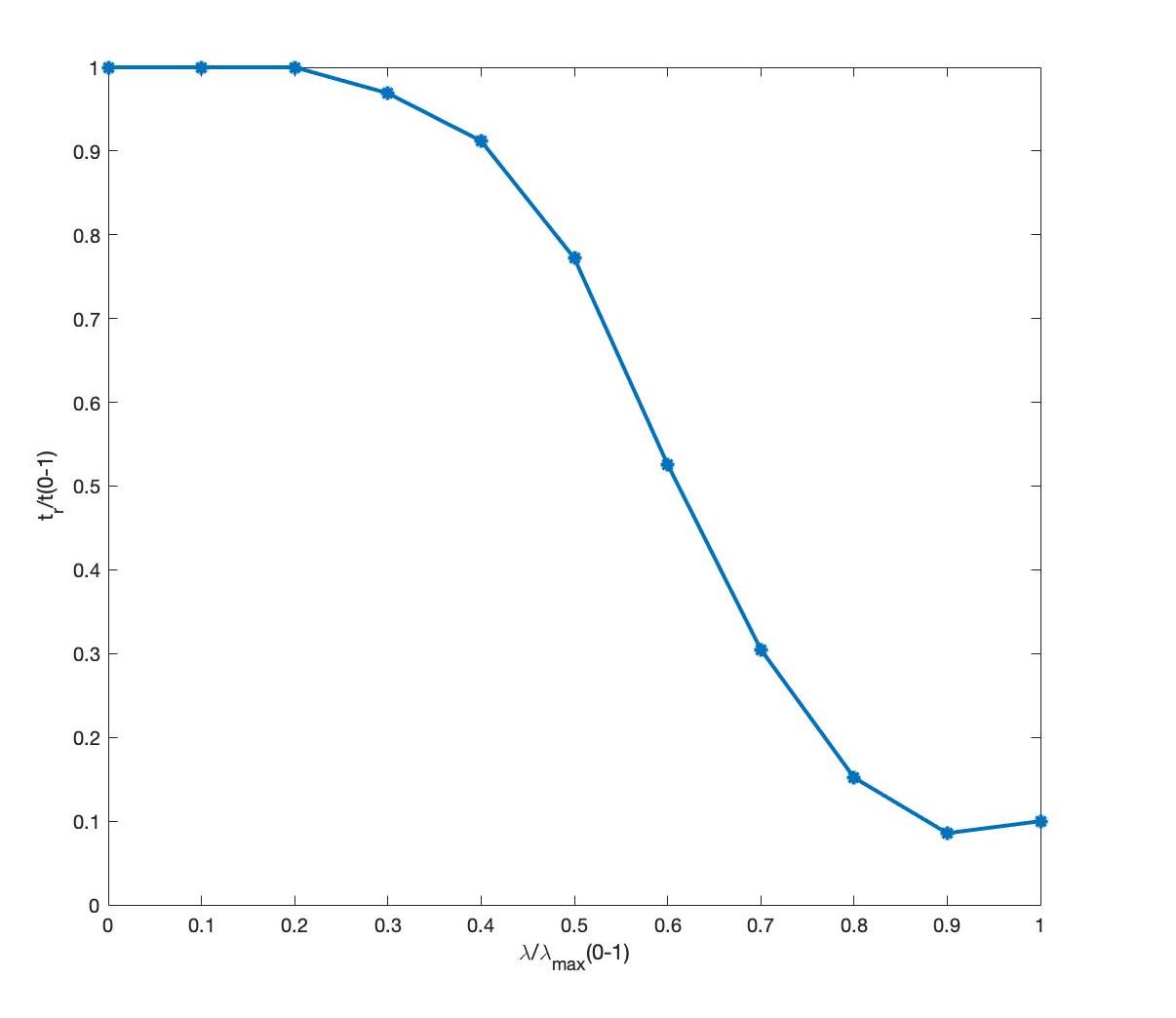}
	\caption{Time reduction - MNIST}
\end{figure}


Moreover, to ensure the optimal solution doesn't change, we can check whether the optimal solution changes by computing:
\begin{align*}
\text{max}\ |\theta_o-\theta_s|
\end{align*}
where $\theta_o$ is the solution without screening, $\theta_s$ is the solution with screening. And the maximum of these absolute values turns to be zero, which indicates the optimal solution doesn't change.

Similarly, the two figures can also be plotted for NYTW as:

\begin{figure}[H]
	\centering
	\includegraphics[width=0.55\linewidth]{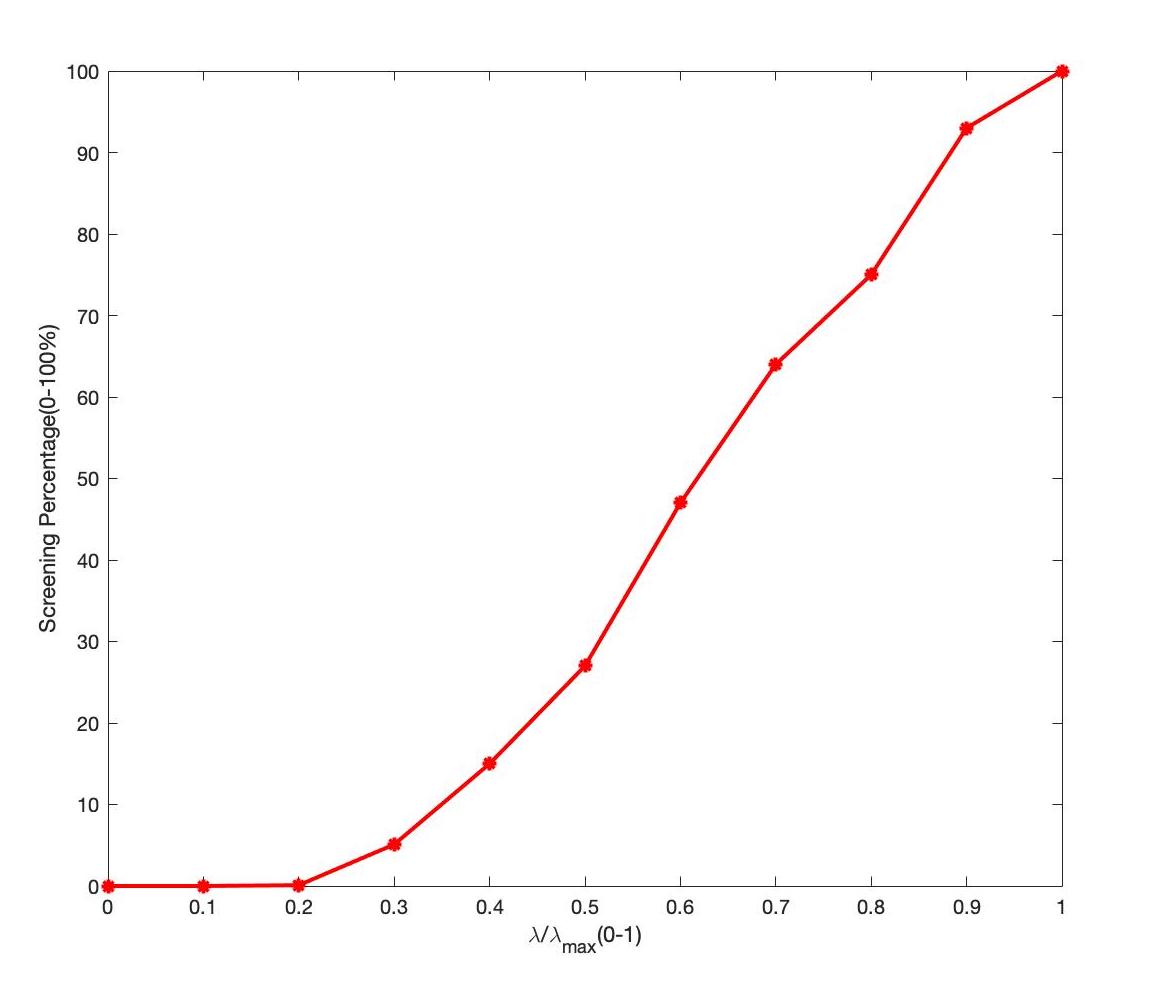}
	\caption{Screening percentage - NYTW}
\end{figure}

\begin{figure}[H]
	\centering
	\includegraphics[width=0.55\linewidth]{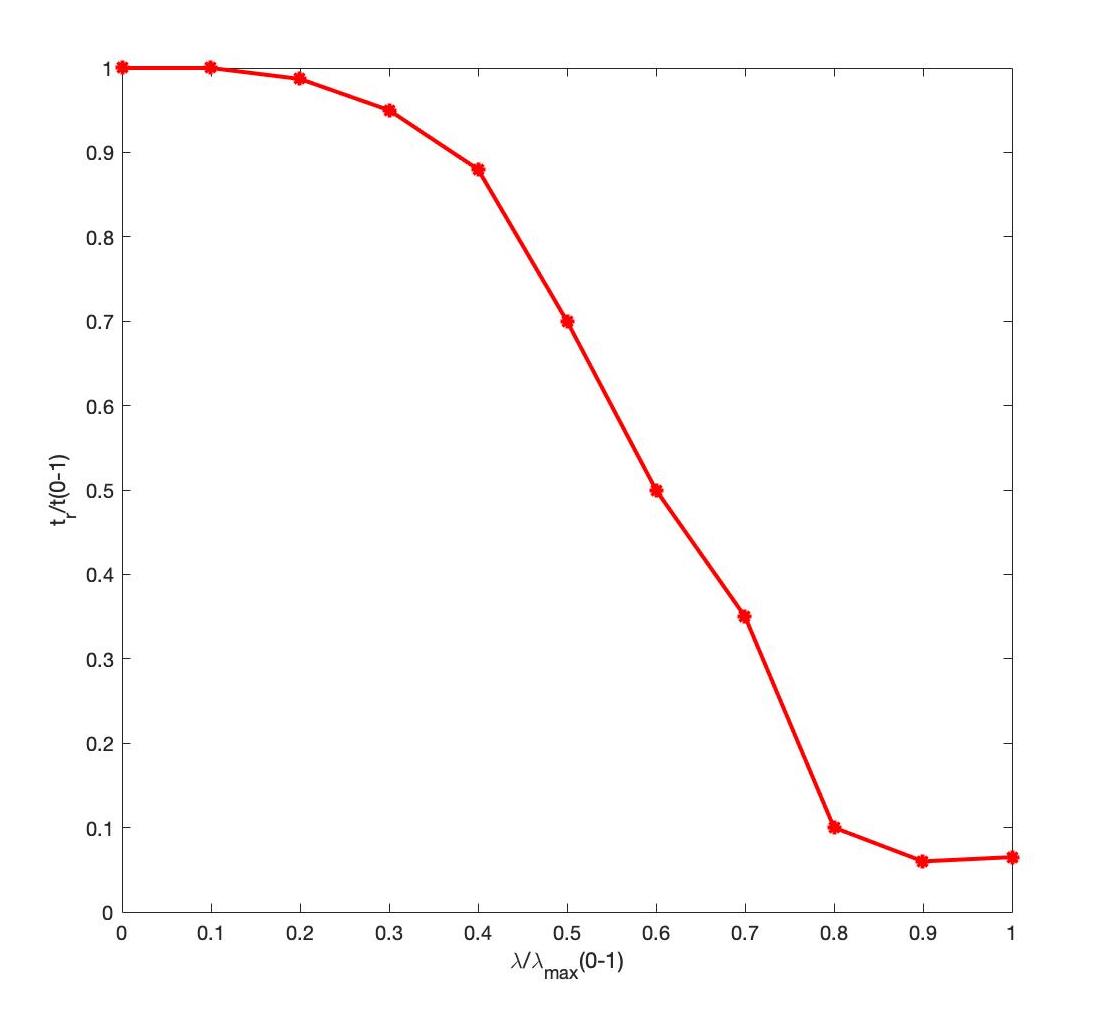}
	\caption{Time reduction - NYTW}
\end{figure}

The two curves are a bit different from those of MNIST, while the tendencies are alike.

\subsubsection{Conclusions}
In this section, we manage to speed up sparse Bayesian learning by screening test. As we can see in the figures, the acceleration will increase as $\lambda/\lambda_{\text{max}}$ goes larger, especially when $\lambda=\lambda_{\text{max}}$, the region $\mathcal{R}$ for the region test is nearly empty, thus almost all the features are rejected, which is is consistent with the our illustration in region test. What's more, to verify the proposed sparse Bayesian learning does work smoothly without making damage to the original optimal solution, we also checked whether the two solutions are identical.

We should note that this acceleration is not so attractive when $\lambda/\lambda_{\text{max}}$ is too small, which is consistent with the performance of the THT in \cite{XiangWR14}, this can be explained by observing the region $\mathcal{R}$. The smaller $\lambda$ is, the larger the sphere will be, thus the looser our bound will become.

What's more, considering what $\lambda$ represents (the noise variance for the linear system), the larger it is, the noisier our system will be. For different data sets, the numerical performances of the proposed sparse Bayesian learning with screening test should be different, however it still can be concluded that the screening test is indeed safe and efficient.

By choosing $\lambda$ appropriately, the optimal solution with respect to the specific $\lambda$ will be obtained more efficiently without making too much damage to the accuracy. In other words, there is a trade-off between acceleration and accuracy.

\section{Application to Classification Problem}\label{chapter05}

In this section, we will apply the proposed method to do classification for real-world data sets. We will do classification for  MNIST\cite{mnist1} data set, which we have used in the last section.


\subsection{Introduction}

In the last section, we had a brief introduction for MNIST, and used it to verify the proposed sparse Bayesian learning with screening test does outperform in speed while keeping the optimal solution unchanged at the same time. However, the simulation in the last section is lacking in value of application, in other words, we only verified that screening test works for sparse Bayesian learning, but ignored the discussion on how the acceleration via screening test can make contributions to real-world applications.

Now we will do classification for MNIST by the proposed method to check its practical performance. The figure below provides some samples in MNIST indicating the images can be classified with respect to the digits $0,1,\ldots,9$:

\begin{figure}[H]
	\centering
	\includegraphics[width=0.6\linewidth]{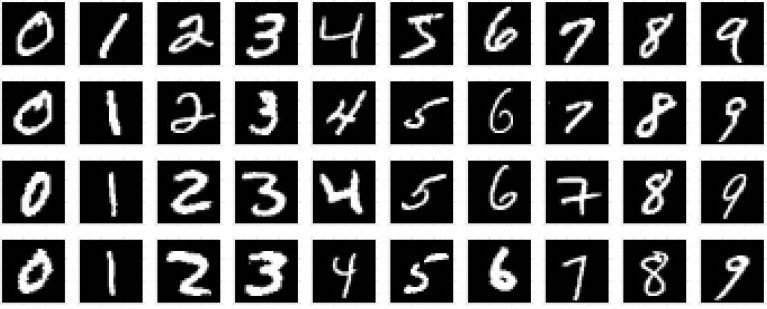}
	\caption{Samples for MNIST}
\end{figure}

This dataset is a popular tutorial for image classification in machine learning, for which lots of techniques and frameworks have been developed. The $70,000$ images ($60,000$ for training and $10,000$ for testing) of handwritten digits are in grayscale and share a resolution of $28\times 28$. What's more, the numerical pixel values for the images are integers between $0$ and $255$.

\subsection{Methodology}
The simulation settings are similar to what we did in Section 4, we vectorize and scale the images in the data set to construct a linear system in (\ref{eq:lm}). However, this time we will do classification by cross validation with respect to the optimal solution obtained for different $\lambda/\lambda_{\text{max}}$.
\bigskip

The methodology is shown as below:

\begin{enumerate}
	\item To make the result more convincing, we will make use of Monte-Carlo method\cite{monte_carlo}, which defines the first loop of size $N_1$.
	\item Next, for each $N_1$, the same grid of $\lambda$ will be generated, the length of grid should be $N_2$, which is our second loop.
	\item For each $N_1$ and the specific grid of $\lambda$, we randomly choose $N_3$ target images as a testing batch for $Y$, and find the sparse representations accordingly by the proposed method based on the $10,000$ images selected in $\Phi$, i.e.:
	\begin{align}
	Y^{(i)} = \Phi\theta^{(i)} + V^{(i)}, i=1,\ldots,N_3
	\end{align}
	where $Y^{(i)}\in\mathbb{R}^{784}$ is the vectorized target image, $\Phi\in\mathbb{R}^{784\times 10000}$, $V^{(i)}\in\mathbb{R}^{784}$ is the unknown noise vector, and $\theta^{(i)}\in\mathbb{R}^{10000}$ is the parameter to be estimated.
	\item Since the columns in the regression matrix $\Phi$ represent different handwritten digits, we can accumulate the elements in $\theta$, i.e., weights of the feature images, to decide the classification. Since the weights could be negative, so we will add up the absolute values of $\theta_i$:
	\begin{align}
	ABS_k = \sum\limits_{\phi_i\text{ represents digit k}}|\theta_i|,  k=0,\ldots,9
	\end{align}
	and then define the metric $prob_k$ as:
	\begin{align}
	prob_k =\frac{ABS_k}{ \sqrt{\sum\limits_{i=1}^k ABS_k^2}}, k=0,\ldots,9
	\end{align}
	where $prob_k\in[0,1]$.
	\item Decide the classification by the largest $prob_k$, and compare it with the truth.
	\item For each value of $\lambda$, we should first gather $N_3$ classification results to obtain the classification accuracy for each Monte-Carlo simulation, and then compute the average accuracy with respect to $N_1$ Monte-Carlo simulations as overall accuracy. The overall accuracy should be with respect to the defined grid of $\lambda$. Standard error of the overall accuracy should be available as well.
\end{enumerate}

\subsection{Simulation Result}

We let $N_1=50,N_2=11,N_3=100$, i.e., the number of Monte-Carlo simulations is $50$, $\lambda$ is selected as $[0,0.1,\ldots,0.9,1.0]$, and $100$ images are considered in the testing batch.

To visualize the prediction, we can make use of color bar to display the value of $prob_k$. For example, we can check the prediction with respect to a small interval of $\lambda$ as below:
\begin{figure}[H]
	\centering
	\includegraphics[width=0.95\linewidth]{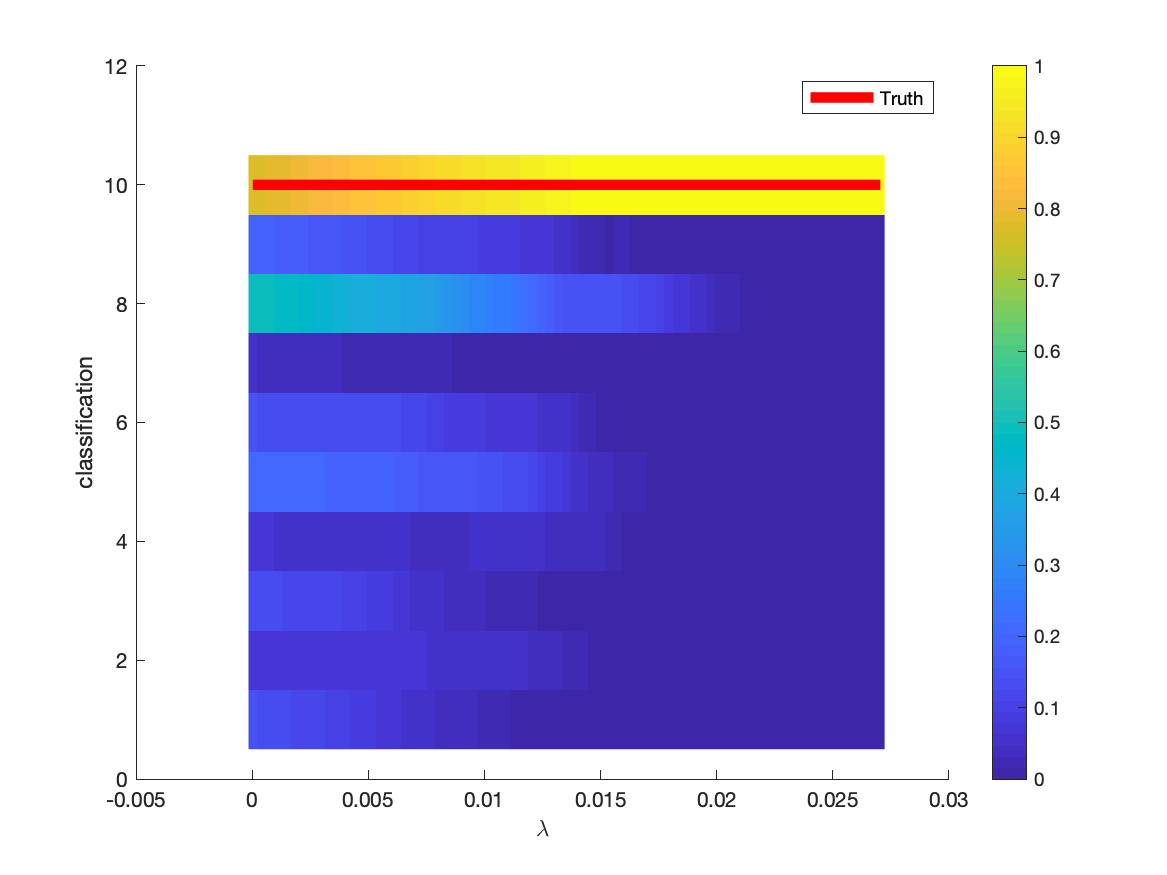}
	\caption{Classification for MNIST - color bar}
	\label{app1_bar}
\end{figure}

As we can see, for a fixed $\lambda$, the red line represents the true digit of the target image, while the color blocks represent the values of $prob_k$, and the colors are decided with respect to the color bar on the right side of the figure. In this figure, as $\lambda$ increases from zero, the prediction will be closer to the truth. However, this is only the case for a small interval of $\lambda$; also, it's just one of the images in the testing batch, the overall accuracy should be computed based on $100$ testing images and $50$ Monte-Carlo simulations.  

Based on all the simulations, finally we can obtain the classification accuracy with standard error as below:

\begin{figure}[H]
	\centering
	\includegraphics[width=0.8\linewidth]{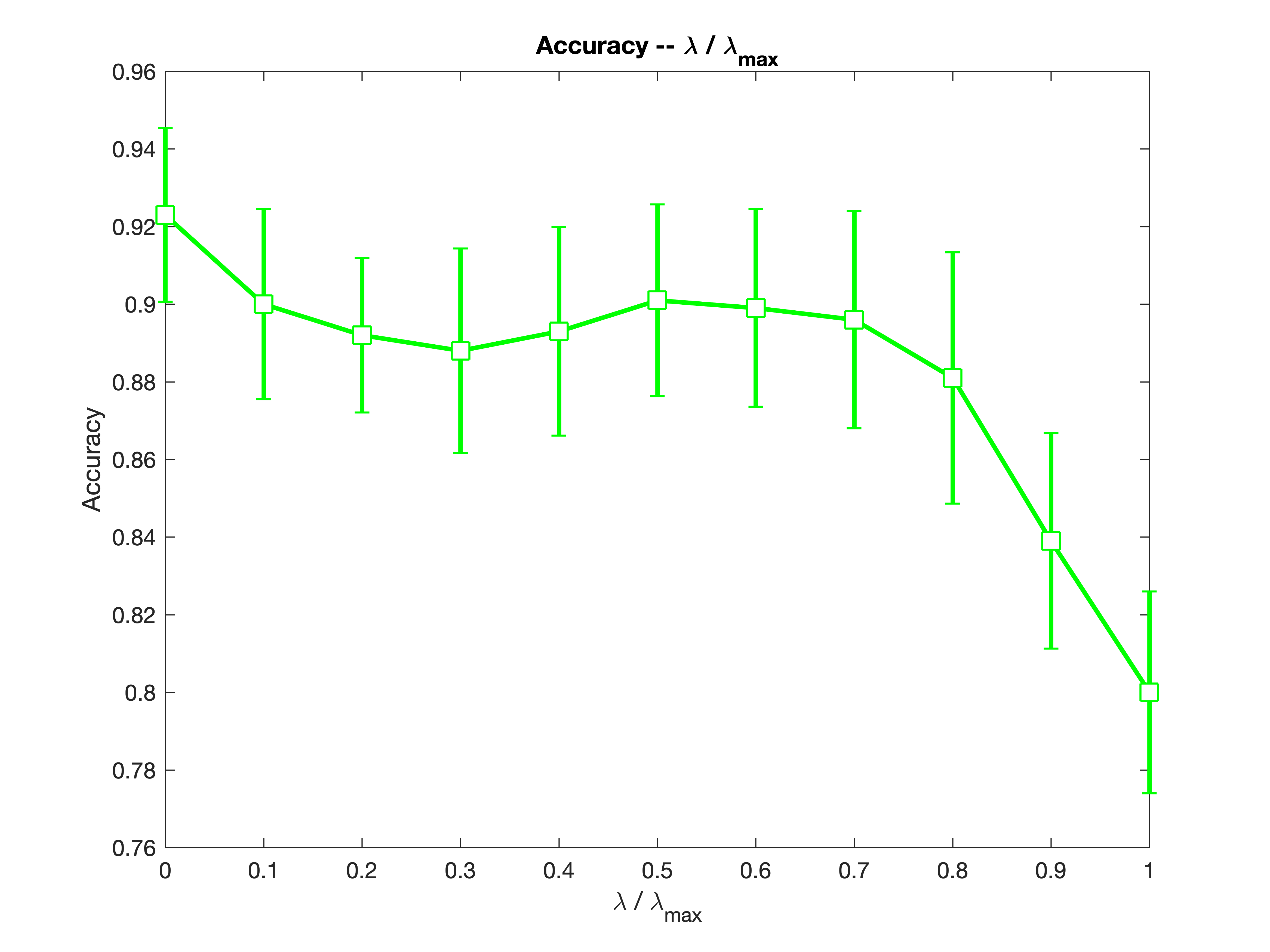}
	\caption{Classification accuracy with standard error - MNIST}
\end{figure}

This figure indicates that as $\lambda/\lambda_{\text{max}}$ goes larger, the accuracy for classification will decrease first, increase afterwards, and decrease again in the end. Even though in this simulation, we obtain the largest accuracy when $\lambda/\lambda_{\text{max}}\in(0,0.1)$, it's still acceptable to sacrifice some accuracy to save computation time.

\subsection{Conclusions}

This section examines the performance of the proposed method on a classical data set for classification: MNIST, where the classification is decided by the scaled accumulation of weights. As Section 3 indicates, the acceleration by screening is not so attractive when $\lambda/\lambda_{\text{max}}$ is too small. So in this application, we have two goals:
\begin{itemize}
	\item To make sure sparse Bayesian learning works for such kind of classification.
	
	This is the minimum requirement, otherwise the acceleration will have no foundation.
	\item To explore whether significant acceleration can be achieved.
	
	Even if sparse Bayesian learning works, we cannot make sure whether to use screening test is meaningful. If the classification accuracy crashes as $\lambda$ goes too large, then the acceleration will be unreasonable. We want to select a $\lambda$ that balances the acceleration and accuracy.
\end{itemize}

The simulation results indicate that our classification for MNIST can achieve both of the two goals successfully.
\section{Application to Signal Reconstruction}\label{chapter06}

In this section, we will apply the proposed method to signal reconstruction in astronomical imaging. In signal reconstruction and image processing, provided with the prior knowledge that the signal (or image) has very few nonzero components, sparse Bayesian learning with screening test can be put into good use.

Astronomical images with many pixels can be represented by a series of point sources. To achieve source localization and denoising, we will model the signal as a linear combination of a set of features. We should also note that this framework is not limited to astronomical imaging, but can also be extended to other systems that can be modeled alike.

\subsection{Problem Formulation}
In this application, the proposed method will be used for performing dictionary learning to determine an optimal feature set for reconstructing a signal representing light sources. The signal of multiple light sources to be constructed should be generated as linear combinations of single-source signals with Gaussian noise, and the performance of reconstruction will be evaluated according to scientific metrics.

First, we should introduce a fluorescence model as described in \cite{app2}. For a single source, the expected photon count depends on the choice of point spread function (PSF). Here we approximate a 3-dimensional PSF by a Gaussian distribution as below:
\begin{equation}
\operatorname{PSF}(x, y, z)=\frac{1}{\sqrt{8 \pi^{3} \sigma_{x y}^{2} \sigma_{z}}} e^{-\frac{1}{2 \sigma_{x y}^{2}}\left[\left(x-x_{0}\right)^{2}+\left(y-y_{0}\right)^{2}\right]-\frac{\left(z-z_{0}\right)^{2}}{2 \sigma_{z}^{2}}}
\end{equation}

Then the PSF must be integrated over the pixel area to become the expected photon count at each pixel:
\begin{equation}\label{app2_form}
\mu_{i j k} =I_{i j k} \Delta E_{x y}\left(x_{i}-x_{0}\right) \Delta E_{x y}\left(y_{j}-y_{0}\right) \Delta E_{z}\left(z_{k}-z_{0}\right)+b g
\end{equation}
with
\begin{equation}
\Delta E_{\mathrm{k}}(u)=\frac{1}{2}\left[\operatorname{erf}\left(\frac{u+\frac{1}{2}}{\sqrt{2} \sigma_{\mathrm{k}}}\right)-\operatorname{erf}\left(\frac{u-\frac{1}{2}}{\sqrt{2} \sigma_{\mathrm{k}}}\right)\right]
\end{equation}
where $I$ is the intensity, $(x_i,y_j,z_k)\in\R^3$ are the pixel coordinates in unit of pixel, $(x_0,y_0,z_0)\in\R^3$ is the location of light source, $bg\in\R$ is the background intensity, $\operatorname{erf}(\cdot)$ is the error function encountered in integrating the normal distribution, and $\sigma_k$ including $\sigma_{x y}$ and $\sigma_z$ are the variances.

While in our application, we will reconstruct a blurred 2-dimensional target image with multiple sources based on a dictionary of single-source images (features), therefore the fluorescence model will degenerate to 2-dimensional accordingly. Then the PSF should be:
\begin{align}
\operatorname{PSF}(x, y)=\frac{1}{\sqrt{4 \pi^{2} \sigma_{x y}^{2} }} e^{-\frac{1}{2 \sigma_{x y}^{2}}\left[\left(x-x_{0}\right)^{2}+\left(y-y_{0}\right)^{2}\right]}
\end{align}
and the expected photon count for pixel $(i,j)$ should be:
\begin{align}
\mu_{i j} =I_{i j} \Delta E_{x y}\left(x_{i}-x_{0}\right) \Delta E_{x y}\left(y_{j}-y_{0}\right) +b g
\end{align}

Then we can generate a target image with $m$ light sources according to the following PSF:
\begin{align}
PSF_{target}(x,y) = \sum_{i=1}^m \theta_i PSF_i(x,y)
\end{align}
where $\theta_i\in\mathbb{R}_+$\footnote{$\theta_i\geq 0$ because intensity cannot be negative.} is the weight of the $i$th single source. An example for a target image with four light sources is shown as below:
\begin{figure}[H]
	\centering
	\includegraphics[width=0.5\linewidth]{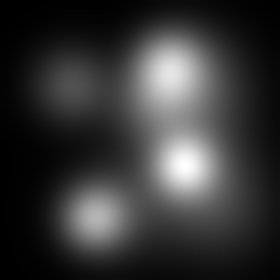}
	\caption{Target image}
	\label{app2_tar}
\end{figure}

As for features, they will be generated with the same resolution of the target image according to a dictionary of single-source PSFs. Four examples for features are listed as below:
\begin{figure}[H]
	\begin{minipage}{0.48\linewidth}
		\centerline{\includegraphics[width=3.0cm]{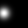}}
		\centerline{(a) Feature 1}
	\end{minipage}
	\hfill
	\begin{minipage}{.48\linewidth}
		\centerline{\includegraphics[width=3.0cm]{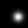}}
		\centerline{(b) Feature 2}
	\end{minipage}
	\vfill
	\begin{minipage}{0.48\linewidth}
		\centerline{\includegraphics[width=3.0cm]{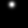}}
		\centerline{(c) Feature 3}
	\end{minipage}
	\hfill
	\begin{minipage}{0.48\linewidth}
		\centerline{\includegraphics[width=3.0cm]{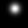}}
		\centerline{(d) Feature 4}
	\end{minipage}
	\caption{Examples of feature images}
	\label{fig:res}
\end{figure}

Note that all the images used will be generated as $28\times 28$\footnote{In this section, the figures to show the performance, including target image, blurred image, and reconstructed image, have been resized to a larger scale by interpolation method in MATLAB for better display.}, thus we can vectorize these images as $784$-dimensional vectors and construct the response $Y$ and regression matrix $\Phi$ in (\ref{eq:lm}) as:
\begin{align}\label{app2_lm}
Y = \sum_{i=1}^n \theta_i \phi_i(x,y,p_i) + V = \Phi(x,y,p)\theta + V
\end{align}
where $Y\in\R^N$ is the blurred target image to be reconstructed, $\Phi=[\phi_1,\ldots,\phi_n]\in\R^{N\times n}$ is the regression matrix made up of $n$ feature images, $(x,y)$ are pairwise coordinates of pixels with respect to the mesh grid based on $[1,2,\ldots,28]$ and $[1,2,\ldots,28]$, $\theta\in\R^n$ is the weights of features, $bg$ will be set to zero for convenience, which means we will generate $Y$ and $\Phi$ under the same background intensity, $V\sim\mathcal{N}(0,\lambda I_N)$ is the noise vector, $\lambda\in\R_+$. As for $p$, we have $p_i=(x_{0_i},y_{0_i},\sigma_{xy_i}),i=1,\ldots,n.$

Then the parameter set to be estimated, say $\Theta$, should be:
\begin{align}\label{param_set}
\Theta=(\theta,p)=(\theta_1,\ldots,\theta_n,p_1,\ldots,p_n)\in\R^{n+3n}
\end{align}

Notice that $\Theta$ can be divided into two parts with respect to being linear to $Y$ or not: $\theta$ is the linear part, while $p$ is the nonlinear part. As discussed in Section 2 and Section 3, the occasions to use the proposed method should satisfy that the parameter to be estimated is linear to the response. Therefore, in the next section, we will try to find a reasonable $p$ by sampling. 

\subsection{Sampling}

In this section, we will decide $p$ by sampling. Sampling is a process used in statistical analysis, in which a specific number of observations are selected from a larger observation pool. Since $p$ includes $x_0,y_0,\sigma_{xy}$, the sampling is equivalent to finding the $n$ triples of parameters $(x_{0_i},y_{0_i},\sigma_{xy_i})$ that define the $n$ features in the regression matrix $\Phi$.

Theoretically, our sampling should be based on the prior of $p$. Even in the worst case where we have no idea how the light sources in the target image are distributed, we can still sample $p$ with respect to Gaussian distribution or uniform distribution. As $n$ goes larger, our samples should be able to cover more possible features, which will definitely influence the performance for reconstruction. In our simulation, we sample $10,000$ features to construct $\Phi$.

Since parameters in $p$ are obtained, multiple pairs of $(x,y)$ representing pixel units are known inputs, thus all features can be generated accordingly with respect to PSF; then we can finally model the problem as:
\begin{align}
Y=\Phi\theta+V
\end{align}
where $Y\in\mathbb{R}^{784}$ is the target vector, $\Phi\in\mathbb{R}^{784\times 10000}$ is the feature matrix, $\theta\in\R^{10000}$ is the parameter to be estimated, $V\in\R^{10000}$ is the noise vector in actual observations, and $V\sim\mathcal{N}(0,\lambda I_{784})$. Now the proposed method is applicable.

Note that unlike the classification in the last section, this time we will introduce the noise $V$ manually. Then the image in Figure \ref{app2_tar} will be blurred as:

\begin{figure}[H]
	\centering
	\includegraphics[width=0.5\linewidth]{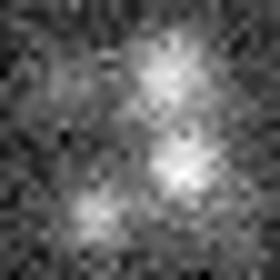}
	\caption{Blurred target image}
	\label{app2_tar_blur}
\end{figure}

For the recovery of the blurred image, we have two goals to achieve:
\begin{itemize}
	\item The first is source localization, which aims at recovering the true light sources in the generation of the target image. If some $\theta_i$ is non-zero, then the corresponding $\phi_i$ will be included in the sparse representation, then the light source center $(x_{0_i},y_{0_i})$ will show up in the reconstructed image. The performance will be evaluated with respect to a self-defined metric.
	
	\item The second is denoising. We want more information besides locations for light sources, which means we hope to recover the entire image efficiently. The performance will be evaluated with respect to a traditional metric and compared with built-in denoising methods in MATLAB.
\end{itemize}

\subsection{Source Localization}

As the title indicates, source localization is the detection of the light sources in an image. After we obtain an optimal $\hat{\theta}$ by solving the linear system, we will be able to reconstruct the target image as:
\begin{align}
\hat{Y}=\Phi\hat{\theta}
\end{align}

If some $\hat{{\theta}}_i$ in $\hat{{\theta}}$ is non-zero, the corresponding feature image $\phi_i$ should be included in the reconstruction, and thus the light source $(x_{0_i},y_{0_i})$ will be detected. To make the simulation more convincing, $20$ different target images will be generated and the statistics will be averaged accordingly. The following figures show the average performance of this application. First, we check the performance of the proposed method with respect to screening percentage:

\begin{figure}[H]
	\centering
	\includegraphics[width=0.6\linewidth]{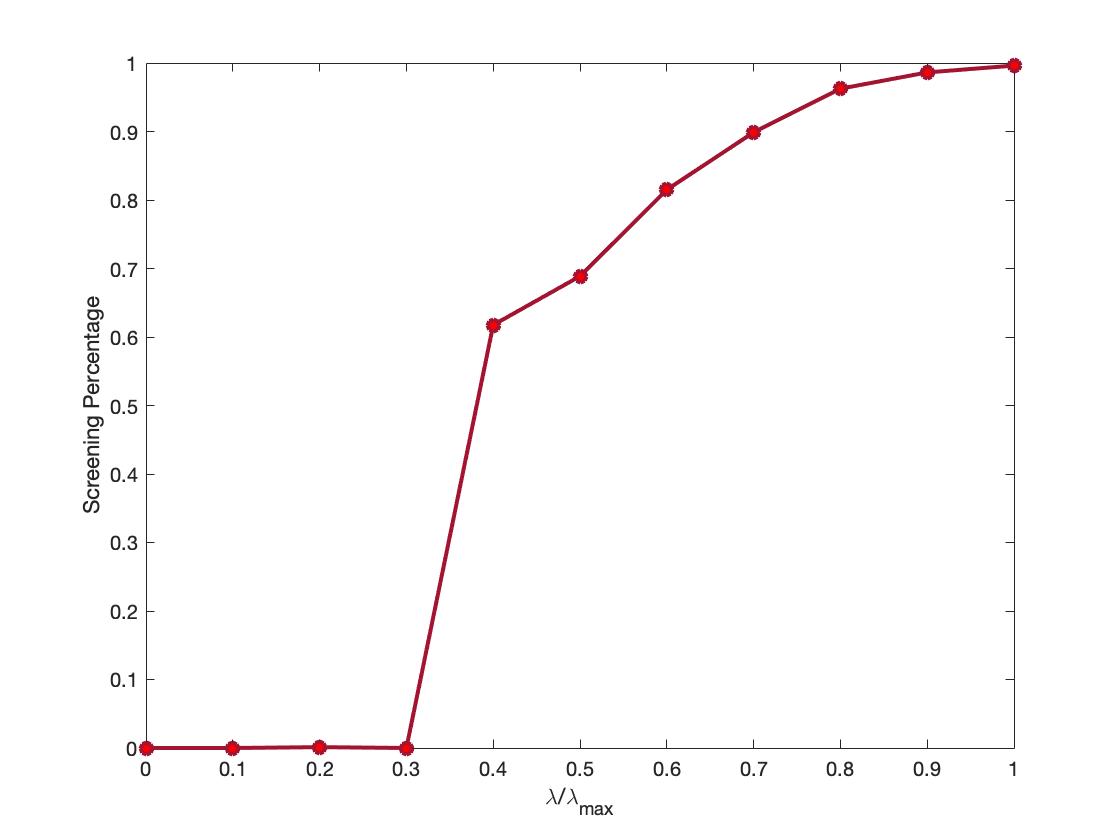}
	\caption{Screening percentage - source localization}
	\label{fig:6.4}
\end{figure}

The figure indicates that the screening percentage increases rapidly as $\lambda/\lambda_{\text{max}}$ becomes larger than 0.3. Next, we check the time reduction:

\begin{figure}[H]
	\centering
	\includegraphics[width=0.6\linewidth]{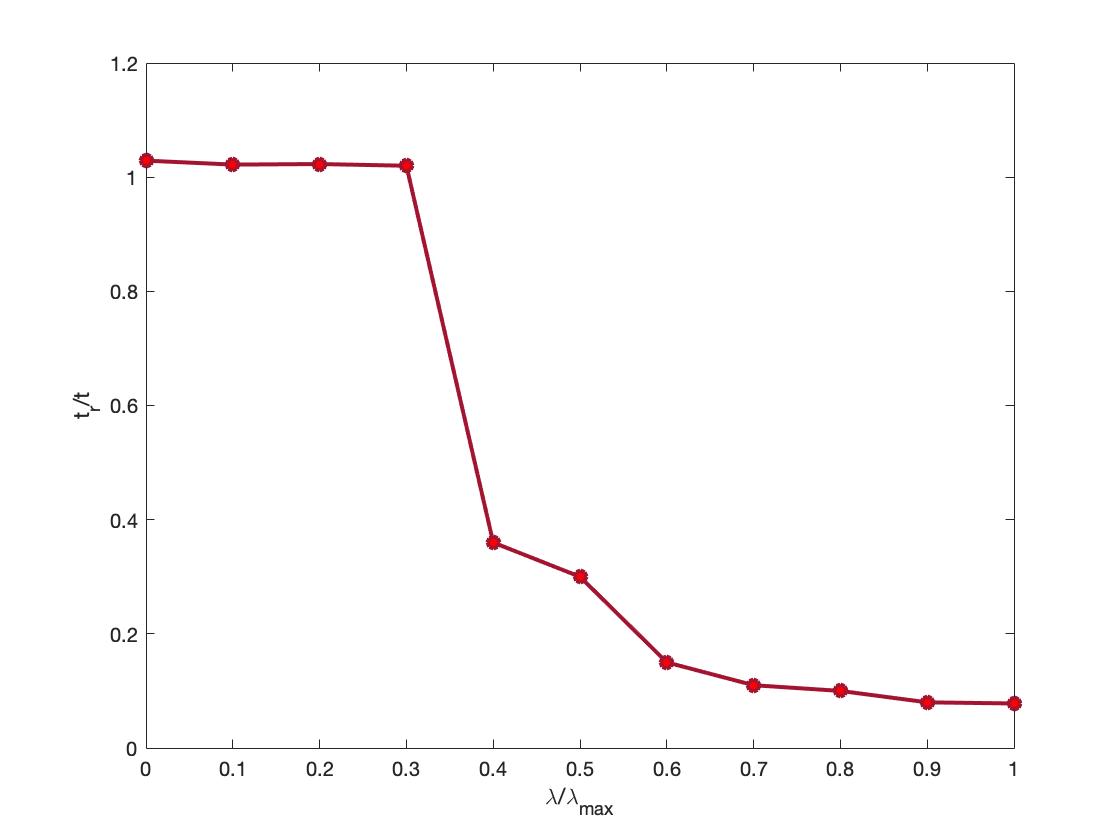}
	\caption{Time reduction - source localization}
\end{figure}

When $\lambda/\lambda_{\text{max}}$ is no larger than 0.3, the reduced time $t_r$ is even larger than the raw time $t$ without screening. Since the screening percentage is too low, little computation time will be saved while the screening will still consume extra time. 

Finally, we observe the whole process that how the sparse solution converges to the true light sources. Four figures are provided below, where the green points are the true sources to be detected, the red circles represent the sparse representation we obtain.

When $\lambda$ is too small and sparsity is not enough:

\begin{figure}[H]
	\centering
	\includegraphics[width=0.5\linewidth]{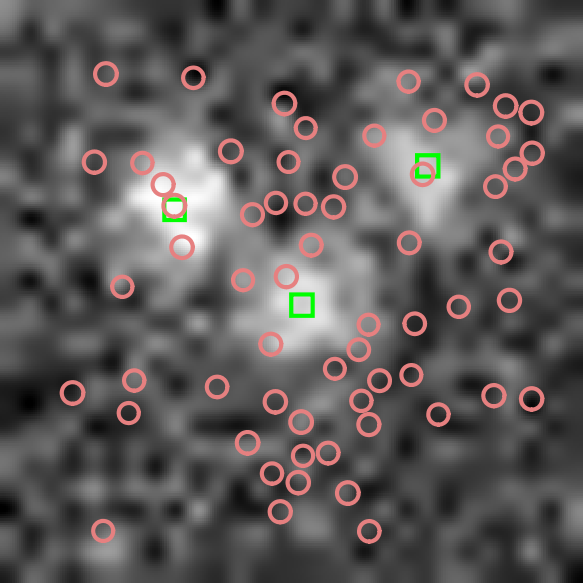}
	\caption{Source localization result - stage 1}
\end{figure}

When $\lambda$ is larger: 

\begin{figure}[H]
	\centering
	\includegraphics[width=0.5\linewidth]{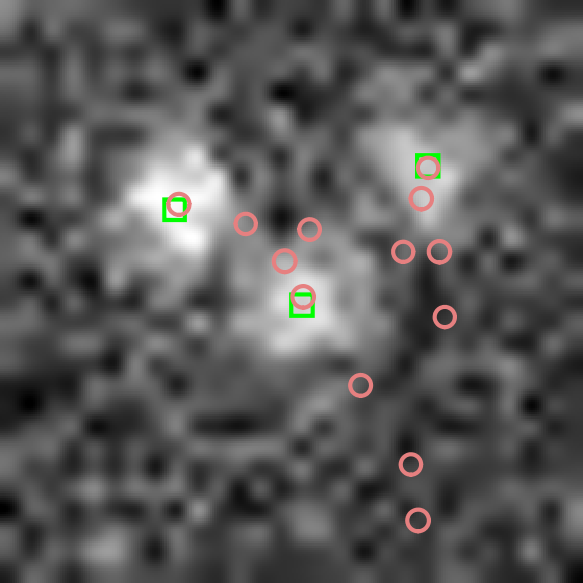}
	\caption{Source localization result - stage 2}
\end{figure}

And $\lambda$ continues increasing:

\begin{figure}[H]
	\centering
	\includegraphics[width=0.5\linewidth]{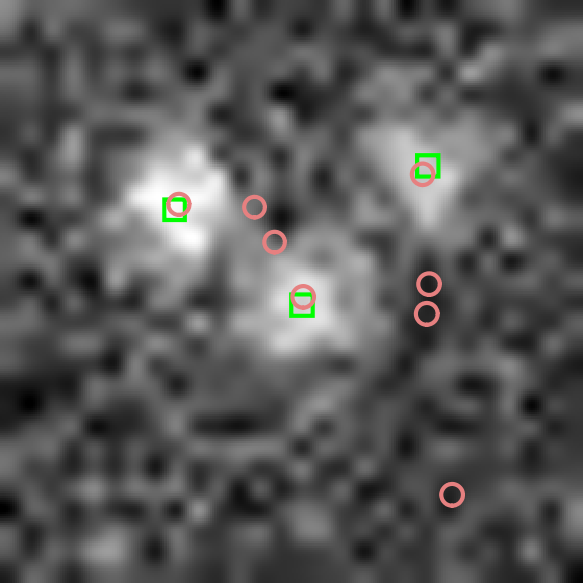}
	\caption{Source localization result - stage 3}
\end{figure}

The most proper $\lambda$ leads to the result below:

\begin{figure}[H]
	\centering
	\includegraphics[width=0.5\linewidth]{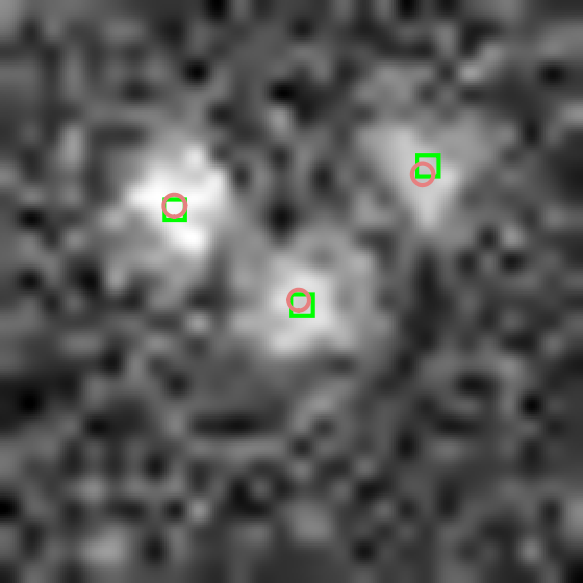}
	\caption{Source localization result - final}
\end{figure}

As we can see, the reconstructed signals based on the sparse representations are gathering around the true light sources gradually as $\lambda$ goes larger, even though the recovery is not completely accurate, it definitely provides us with significant information.

As for the accuracy for the detection, we choose a very popular evaluation metric used in the object detection: intersection over union (IoU)\cite{iou}. IoU, also known as Jaccard Index or Jaccard similarity coefficient, is a statistic used to measure the similarity and diversity of sample sets. It measures similarity between finite sample sets by computing the size of the intersection divided by the size of the union of the sample sets:
\begin{align}
IoU = \frac{DetectionResult\bigcap GroundTruth}{DetectionResult\bigcup GroundTruth}
\end{align}

However, as we can see, the definition of IoU is not enough when the number of detection results and ground truths are different. Therefore, we have to further define group-IoU. In case we mistake some bad detections as good ones, the group-IoU will be defined with respect to $m$ is larger than $n$ or not:  
\begin{Definition}
	Suppose we have $m$ detection results and $n$ ground truths, then:
	\begin{itemize}
		\item When $m> n$, for each detection result, we compute the IoUs between this result and all the ground truths, select the largest one, and then use the average of the $m$ largest IoUs as the group-IoU.
		\item When $m\leq n$, for each ground truth, we compute the IoUs between this truth and all the detection results, select the largest one, and then use the average of the $n$ largest IoUs as the group-IoU.
	\end{itemize}
\end{Definition}

Then we can use this group-IoU as IoU for our detection. It's worth mentioning that in our codes, both the detection result and ground truth are defined as rectangulars in the same size, rather than what is shown in the four figures above. And the IoU for detection can be shown as:
\begin{figure}[H]
	\centering
	\includegraphics[width=0.7\linewidth]{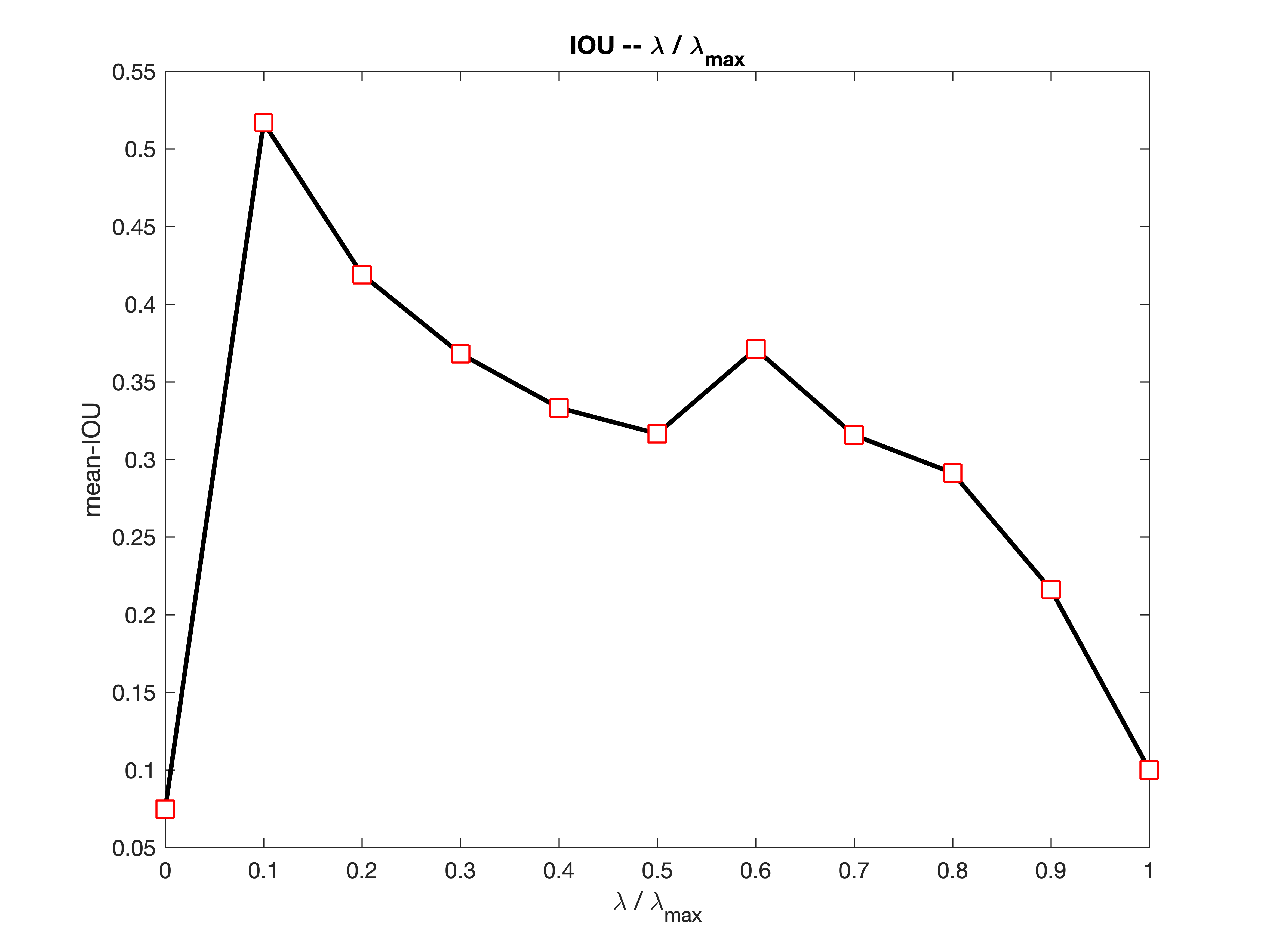}
	\caption{Intersection over union}
	\label{fig:iou}
\end{figure}

We notice that the tendency of IoU curve is more complicated compared with the curves in the previous figures. However, this doesn't mean the characteristics for screening change. As the definition of group-IoU indicates, both numbers and locations of the detection results will influence the value of group-IoU. Therefore, even if we use such an IoU as criterion, the true performance may not be totally decided by IoU. For example, even though Figure \ref{fig:iou} indicates that $\lambda/\lambda_{\text{max}}\in(0.1,0.2)$ guarantees a higher IoU, however when we check the detection results manually, the results for $\lambda/\lambda_{\text{max}}\in(0.5,0.7)$ look a lot better. Thus the defined group-IoU may not be crucial, but it does tell us some significant information.

What's more, unlike the situation in the last section, this time we have numerical information for the noise $V$, therefore it's natural for us to prefer selecting the optimal $\lambda$ as the true variance for noise $V$ in theory; however in practice, it's completely possible that these two may differ.

\subsection{Denoising}

Denoising is the task of removing noise from an image, which leads to our new goal, to pursue the similarity of the original image and the reconstructed image. We will still reconstruct the target image as:
\begin{align}
\hat{Y}=\Phi\hat{\theta}
\end{align}

However, different from source localization, this time we will focus on the similarity between the reconstructed image $\hat{Y}$ and the true image denoted by $Y_0$. The similarity will be quantified by PSNR\cite{psnr} (peak signal-to-noise ratio). The higher the PSNR is, the better our reconstructed image will be. PSNR is defined as below:
\begin{Definition}
	Suppose $I_1$ denotes the matrix data of the original image, $I_2$ denotes the matrix of the reconstructed image; and $m$ represents the number of rows in the images, $n$ represents the number of columns in the images; moreover, $MAX_{I_1}$ is the maximum intensity in our original image, then:
	\begin{align*}
	PSNR = 20 \log _{10}\left(\frac{M A X_{I_1}}{\sqrt{M S E}}\right)
	\end{align*}
	where $MSE=\frac{1}{m n} \sum\limits_{i=1}^{m} \sum\limits_{j=1}^{n}(I_1(i, j)-I_2(i, j))^{2}$.
\end{Definition}


Since the simulation settings are almost identical to source localization, the screening percentage and time reduction for denoising should be the same as well. The only difference is that PSNR will work as a new criterion rather than IoU. The recovery performance for one of the target images is shown as below:
\begin{figure}[htbp]
	\centering
	\begin{minipage}[t]{0.32\linewidth}
		\centering
		\includegraphics[width=2.5cm]{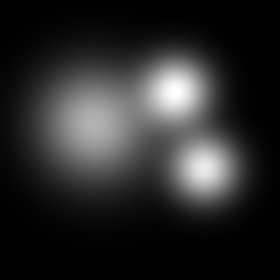}
		\caption{Original}
	\end{minipage}
	\begin{minipage}[t]{0.32\linewidth}
		\centering
		\includegraphics[width=2.5cm]{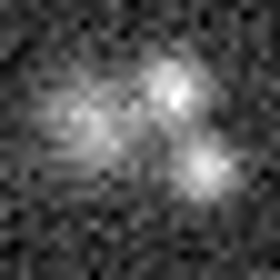}
		\caption{Blurred}
	\end{minipage}
	\begin{minipage}[t]{0.32\linewidth}
		\centering
		\includegraphics[width=2.5cm]{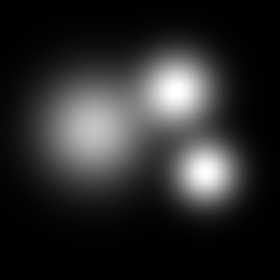}
		\caption{Reconstructed}
	\end{minipage}
\end{figure}

The average accuracy with respect to PSNR is shown as:

\begin{figure}[H]
	\centering
	\includegraphics[width=0.8\linewidth]{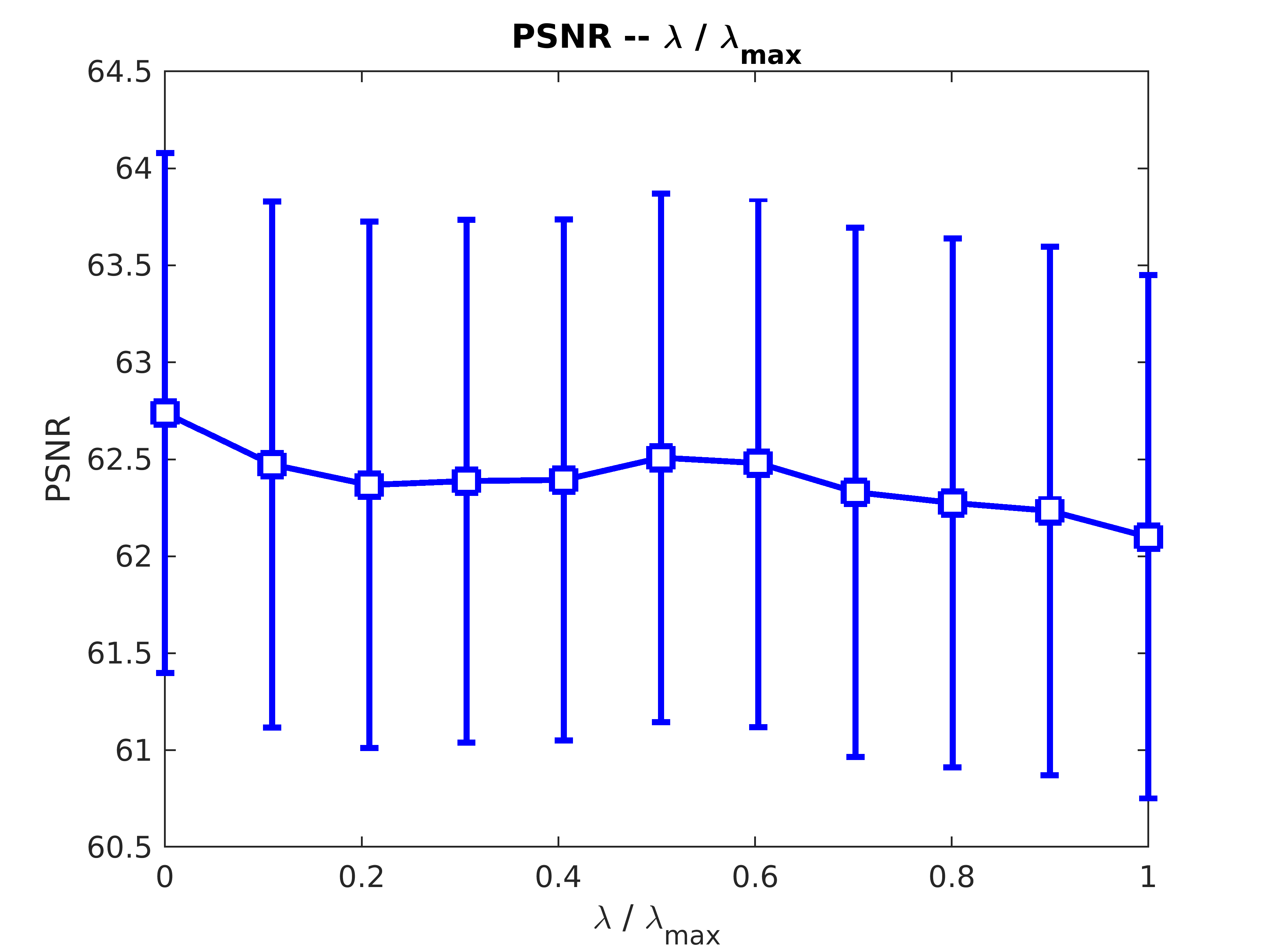}
	\caption{Denoising accuracy - PSNR}
\end{figure}

In both theory and practice, we find that $\lambda/\lambda_{\text{max}}\in (0.5,0.6)$ yields a satisfying performance. Moreover, we compare the performance of the proposed method with traditional algorithms for denoising, for example, wavelet signal denoising method. During the simulation, we generate the data set with different noise variances. When the signal-to-noise ratio (SNR) is small, there's no significant difference between wavelet signal denoising and sparse Bayesian learning with screening test; however, when SNR goes too large, sparse Bayesian learning with screening test will definitely outperforms the wavelet method, which is consistent with our conclusions in Section 3. The figure below shows the performances when SNR=$0.4$.

\begin{figure}[H]\label{fig:group}
	\begin{minipage}{0.48\linewidth}
		\centerline{\includegraphics[width=3.0cm]{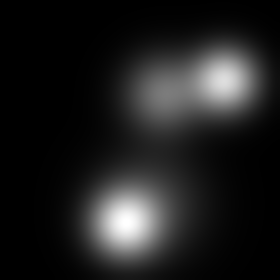}}
		\centerline{Original image}
	\end{minipage}
	\hfill
	\begin{minipage}{.48\linewidth}
		\centerline{\includegraphics[width=3.0cm]{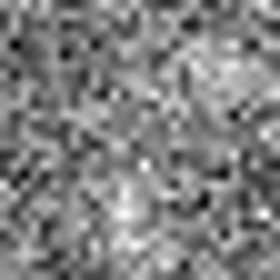}}
		\centerline{Image with Gaussian noise}
	\end{minipage}
	\vfill
	\begin{minipage}{0.48\linewidth}
		\centerline{\includegraphics[width=3.0cm]{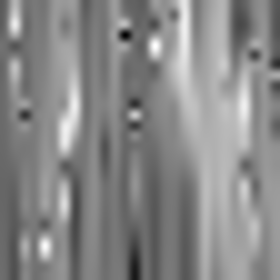}}
		\centerline{Reconstructed - wavelet}
	\end{minipage}
	\hfill
	\begin{minipage}{0.48\linewidth}
		\centerline{\includegraphics[width=3.0cm]{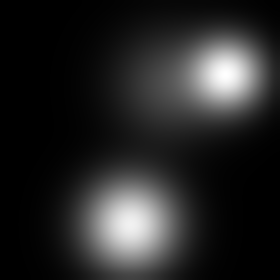}}
		\centerline{Reconstructed - SBL}
	\end{minipage}
	
	\caption{Comparison between SBL with screening test and wavelet denoising under high SNR}
	\label{fig:compare2}
\end{figure}
\subsection{Conclusions}

In this section, the proposed method is applied to signal reconstruction in astronomical imaging. This application has two parts, one is source localization, the other is denoising.

Since the limitations of the proposed method still exist, the two goals mentioned in the conclusions of Section 3 should be inherited. And the simulation results indicate that we achieve both the two goals successfully. Moreover, the reconstruction performs especially well in high-SNR occasions.

What's more, the methodology of this application is obviously more complicated than the application in the last section. That is because even though we manage to model the problem as a linear system, the parameter space $\Theta$ is not completely linear to the response $Y$, thus we have to use sampling as a pretreatment to deal with the non-linear part before using sparse Bayesian learning. Therefore the overall performance will not only rely on our proposed method, but also depend on the pretreatment.

As we said at the beginning of this section, this framework should not be limited to astronomical imaging, but can also be extended to other systems that can be modeled alike.

\section{Conclusions}
As the era of big data is coming, the inter-discipline between traditional statistical methods and machine learning shall draw more and more attention continuously, and the needs for exploration on sparsity will persist as well.

In this work, to find a sparse solution $\theta$ to a linear system more efficiently, we apply screening test to sparse Bayesian learning, thus the new algorithm can inherit the characteristics of sparse Bayesian learning while achieving an acceleration at the same time, which indicates its potential to influence related fields.

In Section 2 and Section 3, we introduce the methodology of sparse Bayesian learning and design a screening test for it, then we examine the performance on two real-world data sets. Though the simulation shows a fairly good performance, we should admit some limitations of the proposed method listed as follows:
\begin{itemize}
	\item The proposed method only works on sparse Bayesian learning that is equivalent to a weighted $\ell_1$-minimization problem, but cannot be used for all types of sparse Bayesian learning.
	
	\item According to the methodology, whether an efficient bound for $\hat{\eta}$ is chosen will definitely influence the performance, thus we have to admit that, both in theory and practice, the performance of Algorithm \ref{alg_wtht} is no better than that of the THT in \cite{XiangWR14}, though it has extra advantages of SBL.
	
	\item Last but not least, both the screening ratios of THT and Algorithm \ref{alg_wtht} depend on $\lambda$ too much. The dependency cannot be totally eliminated in theory, however according to the simulation, to obtain a satisfying acceleration, the value of $\frac{\lambda}{\lambda_\text{max}}$ should be no smaller than $0.4$; considering what $\lambda_\text{max}$ represents, this value range of $\lambda$ may not be always acceptable.
\end{itemize}

In Section 4 and Section 5, we examine sparse Bayesian learning with screening test on two applications. One is classification, the other is signal reconstruction (source localization and denoisng). In these applications, we achieve our goals successfully and efficiently. Especially in the second application, we make it to formulate the problem as a linear system, even though the linear relationship does not hold with respect to the full parameter space $\Theta$. For such issue, we choose to estimate the nonlinear parameters by tricks like sampling. Consequently, we must be aware that the overall performance is decided not only by sparse Bayesian learning with screening test, but also the trick we use before sparse Bayesian learning. For example, the accuracy of sampling will definitely impact on the performance of reconstruction. 


\end{document}